\documentclass{tlp}

\usepackage{url}
\usepackage{hyperref}
\usepackage{caption}
\usepackage{amsmath}
\usepackage{mathtools}
\usepackage{booktabs}
\usepackage{todonotes}
\usepackage{comment}

\usepackage{adjustbox}

\usepackage{listings}
\usepackage{subcaption}

\usepackage{algorithm}
\usepackage{algpseudocode}
\usepackage{amsfonts}

\usepackage{tkz-graph}

\usepackage{pgfplots}

\usepackage{tikz}
\usetikzlibrary{decorations.pathmorphing,arrows.meta}
\tikzstyle{node}=[fill=white, draw=black, shape=rectangle, minimum size = 0.7cm]
\tikzstyle{circ}=[fill=white, draw=black, shape=circle, minimum size = 0.4cm]

\newcommand{\upperprob}{\overline{P}}
\newcommand{\lowerprob}{\underline{P}}

\newcommand{\impl}{\ {:\!-}\  }

\DeclareMathOperator*{\argmax}{arg\,max}

\usepackage{xcolor}
\newcommand{\red}[1]{\textcolor{black}{#1}}

\DeclareMathOperator{\lpmln}{\mathrm{LP}^\mathrm{{MLN}}}

\DeclareMathOperator{\program}{\mathcal{P}}

\DeclareMathOperator{\decision}{\mathit{decision}}

\DeclareMathOperator{\utility}{\mathit{utility}}
\DeclareMathOperator{\Util}{\mathit{Util}}
\DeclareMathOperator{\upperutil}{\overline{\Util}}
\DeclareMathOperator{\lowerutil}{\underline{\Util}}

\newcommand{\outX}{\mathbf{X}_O}
\newcommand{\midX}{\mathbf{X}_M}
\newcommand{\innX}{\mathbf{X}_I}

\definecolor{red_plot}{HTML}{F30522}
\definecolor{orange_plot}{HTML}{FA5F22}
\definecolor{yellow_plot}{HTML}{DED712}
\definecolor{pink_plot}{HTML}{FFD0B4}
\definecolor{light_blue_plot}{HTML}{D1E5F0}
\definecolor{blue_plot}{HTML}{67A9CF}
\definecolor{dark_blue_plot}{HTML}{2166AC}
\definecolor{green_plot}{HTML}{20D525}

\newcommand{\textwidthsubfigure}{0.40}
\newcommand{\textwidthresizebox}{1}

\newtheorem{example}{Example}
\newtheorem{definition}{Definition}

\begin{document}

\lefttitle{Cambridge Author}

\jnlPage{1}{26}
\jnlDoiYr{2024}
\doival{10.1017/xxxxx}

\title[Solving Decision Theory Problems with Probabilistic Answer Set Programming]{Solving Decision Theory Problems with Probabilistic Answer Set Programming}

\begin{authgrp}
\author{\gn{Damiano} \sn{Azzolini}}
\affiliation{Department of Environmental and Prevention Sciences -- University of Ferrara, Ferrara, Italy \\
\email{damiano.azzolini@unife.it}}
\author{\gn{Elena} \sn{Bellodi}}
\affiliation{Department of Engineering -- University of Ferrara, Ferrara, Italy \\
\email{elena.bellodi@unife.it}}
\author{\gn{Rafael} \sn{Kiesel}}
\affiliation{TU Wien, Wien, Austria \\ \email{rafael.kiesel@web.de}}
\author{\gn{Fabrizio} \sn{Riguzzi}}
\affiliation{Department of Mathematics and Computer Science -- University of Ferrara, Ferrara, Italy\\
\email{fabrizio.riguzzi@unife.it}}
\end{authgrp}

\maketitle

\history{\sub{xx xx xxxx;} \rev{xx xx xxxx;} \acc{xx xx xxxx}}

\begin{abstract}
  Solving a decision theory problem usually involves finding the actions, among a set of possible ones, which optimize the expected reward, possibly accounting for the uncertainty of the environment.
  In this paper, we introduce the possibility to encode decision theory problems with Probabilistic Answer Set Programming under the credal semantics via decision atoms and utility attributes.
  To solve the task we propose an algorithm based on three layers of Algebraic Model Counting, that we test on several synthetic datasets against an algorithm that adopts answer set enumeration.
  Empirical results show that our algorithm can manage non trivial instances of programs in a reasonable amount of time.
  Under consideration in Theory and Practice of Logic Programming (TPLP).
\end{abstract}

\begin{keywords}
  Probabilistic Answer Set Programming, Decision Theory, Statistical Relational Artificial Intelligence, Credal Semantics, Algebraic Model Counting.
\end{keywords}

\maketitle

\section{Introduction}
\label{sec:intro}
Logic-based languages are well-suited to model complex domains, since they allow the representation of complex relations among the involved objects.
\textit{Probabilistic} logic languages, such as ProbLog~\citep{DBLP:conf/ijcai/RaedtKT07} and Probabilistic Answer Set Programming under the credal semantics~\citep{cozman2020pasp}, represent uncertain data with \textit{probabilistic facts}~\citep{DBLP:conf/iclp/Sato95}, i.e., facts with an associated probability.
The former is based on the Prolog language~\citep{Lloyd87}, while the latter adopts Answer Set Programming (ASP)~\citep{brewka2011asp}.
ASP has been proved effective in representing hard combinatorial tasks, thanks to expressive constructs such as aggregates~\citep{alviano2018aggregates}.
Moreover, several extensions have been proposed for managing uncertainty also through weighted rules~\citep{DBLP:conf/kr/LeeW16}, increasing even further the possible application scenarios.

Usually, decision theory (DT) problems are composed of a set of possible actions, a selection of which define a strategy, and a set of utility attributes, that indicate the utility (i.e., a reward, possibly negative) of completing a particular task.
The goal is to find the strategy that optimizes the overall expected utility.
Expressing DT problems with (probabilistic) logic languages allows the user to find the best action to take in uncertain complex domains.
While DTProbLog~\citep{DBLP:conf/aaai/BroeckTOR10} is a ProbLog extension that solves DT tasks represented with a ProbLog program, no tool, to the best of our knowledge, is available to solve them with a probabilistic answer set language.
We believe that a (probabilistic) ASP-based tool, by providing expressive syntactic constructs such as aggregates and choice rules, would be of great support in complex environments.  
For example, we may be interested in modelling a viral marketing scenario, where the goal is to select a set of people to target with specific ads to maximize sales.
In this scenario, uncertainty could come from the shopping behavior of individuals.

In this paper we close this gap and introduce decision theory problems in Probabilistic Answer Set Programming under the credal semantics (DTPASP).
In particular, we extend Probabilistic Answer Set Programming under the credal semantics (PASP) with \textit{decision atoms} and \textit{utility attributes}~\citep{DBLP:conf/aaai/BroeckTOR10}.
Every subset of decision atoms defines a different strategy, i.e., a different set of actions that can be performed in the domain of interest.
In the viral marketing example, the decisions are whether to target an individual with an ad.
However, there is uncertainty on the actual effectiveness of the targeting action.
At the same time, targeting a person involves a cost that can be represented with a utility attribute with a negative reward.
The fact that a person buys a product, instead, is associated with a positive reward.
Moreover, there can be other factors to consider, such as preferences among the different possible items that can be bought which can be conveniently represented using ASP.

In PASP, since each possible world can have multiple models, queries are associated with probability ranges indicated by a lower and an upper probability,  instead of point values.
Similarly, in DTPASP, we need to consider lower and upper rewards, and look for the strategies that maximize either of the two.
So a solution of a decision theory problem expressed with DTPASP is composed of two strategies.

We \red{have} developed two algorithms to solve the task of finding the strategy that maximizes the lower expected reward and the strategy that maximizes the upper expected reward, together with the values of these rewards.
A first algorithm is based on answer sets enumeration.
This can be adopted only on small domains, so we consider it as a baseline.
We propose a second algorithm, based on three layers of Algebraic Model Counting (AMC)~\citep{10.1016/j.jal.2016.11.031} that adopts knowledge compilation~\citep{DBLP:journals/jair/DarwicheM02} to speed up the inference process. 
Empirical results show that the latter algorithm is significantly faster than the one based on enumeration, and can handle domains of non trivial size.

The paper is structured as follows: in Section~\ref{sec:background} we briefly discuss background knowledge.
Section~\ref{sec:dtproblog} discusses DTProbLog, a framework to solve decision theory problems in Probabilistic Logic Programming.
Section~\ref{sec:2amc} illustrates the task of Second Level Algebraic Model Counting.
Section~\ref{sec:dt_pasp} introduces DTPASP together with the optimization task to solve.
In Section~\ref{sec:algorithm} we describe the algorithms to solve the task that we test against the baseline in Section~\ref{sec:experiments}.
Section~\ref{sec:related} surveys some of the related works and Section~\ref{sec:conclusions} concludes the paper.

\section{Background}
\label{sec:background}
This section introduces the basic concepts of Answer Set Programming and Probabilistic Answer Set Programming.

\subsection{Answer Set Programming}
\label{subsec:asp}
In the following, we will use the verbatim font to denote code that can be executed with a standard ASP solver.
Here we consider a subset of ASP~\citep{brewka2011asp}.
An ASP program (or simply an ASP) is a finite set of disjunctive rules.
A \textit{disjunctive rule} (or simply rule) is of the form
$$
\underbrace{h_1 ; \dots ; h_m}_{\text{head}} \impl \underbrace{b_1, \dots b_n}_{\text{body}}.
$$
where every $h_i$ is an atom and every $b_i$ is a literal.
We consider only \textit{safe rules}, where every variable in the head also appears in a positive literal in the body.
\red{This is a standard requirement in ASP.}
If the head is empty, the rule is called a \textit{constraint}, if the body is empty and there is only one atom in the head, the rule is called a \textit{fact}, and if there is only one atom in the head with one or more literals in the body the rule is called \textit{normal}.
A \textit{choice rule} is of the form $0\{a\}1 \impl b_1,\dots,b_n$ and indicates that the atom $a$ can be selected or not if the body is true.
Usually, we will omit 0 and 1 and consider them implicit.
ASP allows the use of \textit{aggregate} atoms~\citep{alviano2018aggregates} in the body.
We consider aggregates of the form $\# \varphi \{\epsilon_0 ; \dots ; \epsilon_n\} \ \delta g$ where $g$ is called \textit{guard} and can be either a constant or a variable, $\delta$ is a comparison arithmetic operator, $\varphi$ is an aggregate function symbol, and $\epsilon_0, \dots, \epsilon_n$ is a set of expressions where each $\epsilon_i$ has the form $t_1, \dots, t_n : F$ and each $t_i$ is a term whose variables appear in the conjunction of literals $F$.
An example of aggregate atom is $\#count\{A : p(A)\} = 2$, that is true if the number of ground substitutions for $A$ that make $p(A)$ true is $2$.
Here, $2$ is the guard, and $\#count$ is the aggregate function symbol.
The primal graph of a ground answer set program $\mathcal{P}$ is such that there is one vertex for each atom appearing in $\mathcal{P}$ and there is an undirected edge between two vertices if the corresponding atoms appear simultaneously in at least one rule of $\mathcal{P}$.

The semantics of ASP is based on the concept of \textit{answer set}, also called \textit{stable model}.
With $B_\mathcal{P}$ we denote the Herbrand base of an answer set program $\mathcal{P}$, i.e., the set of ground atoms that can be constructed with the symbols in $\mathcal{P}$.
A variable is called \textit{local} to an aggregate if it appears only in the considered aggregate; if instead it occurs in at least one literal not involved in aggregations, it is called \textit{global}.
The grounding of a rule with aggregates proceeds in two steps, by first replacing global variables with ground terms, and then replacing local variables appearing in aggregates with ground terms. 
An interpretation $I$ of $\mathcal{P}$ is a subset of $B_\mathcal{P}$.
An aggregate is true in an interpretation $I$ if the evaluation of the aggregate function under $I$ satisfies the guards. 
An interpretation satisfies a ground rule if at least one of the $h_i$s is true in it when all the $b_i$s are true in it.
A \textit{model} of $\mathcal{P}$ is an interpretation that satisfies all the groundings of all the rules of $\mathcal{P}$.
Given a ground program $\mathcal{P}_g$ and an interpretation $I$, by removing from $\mathcal{P}_g$ of $\mathcal{P}$ all the rules where at least one of the $b_i$s is false in an interpretation $I$ we get the \textit{reduct}~\citep{faber2004recursive} of $\mathcal{P}_g$ with respect to $I$.
An \textit{answer set} (or stable model) for a program $\mathcal{P}$ is an interpretation that is a minimal (under set inclusion) model of the reduct of the grounding $\mathcal{P}_g$ of $\mathcal{P}$.
We indicate with $AS(\mathcal{P})$ the set of all the answer sets of a program $\mathcal{P}$.
An ASP usually has multiple answer sets.
However, we may be interested only in  the \textit{projective solutions}~\citep{gebser2009projective} onto a set of ground atoms $B$, which are given by the set $AS_{B}(\mathcal{P}) = \{A \cap B \mid A \in AS(\program)\}$.

To clarify these concepts, consider the following example:
\begin{example}
\label{ex:asp_dummy}
The following program $\mathcal{P}$ has, respectively, two choice rules, a normal rule, and a disjunctive rule.
\begin{lstlisting}
{a}.
{b}.
qr :- a. 
qr ; nqr:- b.
\end{lstlisting}
The program has 5 answer sets: $AS(\mathcal{P}) = \{\{\}, \{a, qr\}, \{b, qr\}, \{b, qr, a\}, \{b, nqr\}\}$.
If we project the solutions onto the $qr$ and $a$ atoms we get 3 answer sets: $\{\{\}, \{a, qr\}, \{qr\}\}$.
\end{example}

\subsection{Probabilistic Answer Set Programming (PASP)}
\label{subec:pasp}
PASP extends ASP by representing uncertainty with weights~\citep{DBLP:conf/kr/LeeW16} or probabilities~\citep{DBLP:conf/ilp/CozmanM16} associated with facts.
Here we consider PASP under the credal semantics (CS)~\citep{DBLP:conf/ilp/CozmanM16}.
We will use the acronym PASP to also denote a probabilistic answer set program, the intended meaning will be clear from the context.

PASP allows probabilistic facts of the form~\citep{DBLP:conf/ijcai/RaedtKT07} $\Pi_i::f_i$ where $\Pi_i \in [0,1]$ and $f_i$ is an atom.
We only consider ground probabilistic facts.
Moreover, we require that probabilistic facts cannot appear in the head of rules, a property called \textit{disjoint condition}~\citep{DBLP:conf/iclp/Sato95}.
Every possible subset of probabilistic facts (there are $2^n$ of them, where $n$ is the number of probabilistic facts) identifies a \textit{world} $w$, i.e., an ASP obtained by adding the atom of the selected probabilistic facts to the rules of the program.
Each world $w$ is assigned a probability computed as
\begin{equation}
\label{eq:world_probability}
P(w) = \prod_{f_i \in w} \Pi_i \cdot \prod_{f_i \not \in w} (1-\Pi_i).
\end{equation}
With this setting, we have two levels to consider: at the first level, we need to consider the worlds, each with an associated probability.
At the second level, for each world we have one or more answer sets.
Since a world may have more than one model, in order to assign a probability to queries we need to decide how the probability mass of the world is distributed on its answer sets.
We can choose a particular distribution for the answer sets of a world, such as a uniform~\citep{totis_de_raedt_kimmig_2023} or a distribution that maximizes the entropy~\citep{10.5555/3031748.3031798}.
However, here we follow the more general path of the credal semantics and we refrain from assuming a certain distribution for the answer sets of a world.
This implies that queries are associated with probability ranges instead of point values.
Under this semantics, the probability of a \textit{query} $q$ (i.e., a conjunction of ground literals), $P(q)$, is associated with a probability range $[\lowerprob(q),\upperprob(q)]$ where the lower and upper bounds are computed as:
\begin{equation}
\label{eq:lower_upper_prob}
\lowerprob(q) = \sum_{w_i \mid \forall m \in AS(w_i), \ m \models q} P(w_i),\ 
\upperprob(q) = \sum_{w_i \mid \exists m \in AS(w_i), \ m \models q} P(w_i).
\end{equation}
In other words, the lower bound (or lower probability) $\lowerprob(q)$ is given by the sum of the probabilities of the worlds where the query is true in all answer sets and the upper bound (or upper probability) $\upperprob(q)$ is given by the sum of the probabilities of the worlds where the query is true in at least one answer set.
That is, a world contributes to both the lower and upper probability if the query is true in all of its answer sets while it contributes only to the upper probability if the query is true only in some of the answer sets.
Note that $\lowerprob(q) = 1 - \upperprob(not \ q)$ and $\upperprob(q) = 1 - \lowerprob(not \ q)$: \red{this is true if every world has at least one answer set, otherwise inconsistencies must be managed (see the discussion below).}
To clarify this, consider the following illustrative example:
\begin{example}
\label{ex:example_pasp}
The following program has two probabilistic facts, $a$ and $b$, with probabilities of 0.3 and 0.4, respectively.
\begin{lstlisting}
0.3::a.
0.4::b.
qr :- a. 
qr ; nqr:- b.
\end{lstlisting}
There are $2^2 = 4$ worlds, listed in Table~\ref{tab:worlds_ex_dummy}.
Consider the query $qr$.
Call $w_0$ the world where both $a$ and $b$ are false, $w_1$ the world where $a$ is true and $b$ is false, $w_2$ the world where $b$ is true and $a$ is false, and $w_3$ the world where both $a$ and $b$ are true.
$P(w_0) = (1 - 0.3) \cdot (1 - 0.4) = 0.42$, $P(w_1) = 0.3 \cdot (1-0.4) = 0.18$, $P(w_2) = (1-0.3) \cdot 0.4 = 0.28$, and $P(w_3) = 0.3 \cdot 0.4 = 0.12$.
Note that the sum of the probabilities of the worlds equals 1.
For $w_0$, $AS(w_0) = \{\{\}\}$, the query is false (not present) in the answer set $\{\}$ so we do not have a contribution to any of the probabilities.
For $w_1$, $AS(w_1) = \{\{a,qr\}\}$, the query is true in the only answer set, so we have a contribution of $P(w_1)$ to both the lower and upper probability.
For $w_2$, $AS(w_2) = \{\{b,qr\}, \{b,nqr\}\}$, the query is true in only one of the two answer sets, so we have a contribution of $P(w_2)$ only to the upper probability.
For $w_3$, $AS(w_3) = \{\{a,b,qr\}\}$, the query is true in all the answer sets (there is only one) and we have a contribution of $P(w_3)$ to both the lower and upper probability.
Overall, $P(qr) = [P(w_1) + P(w_3), P(w_1) + P(w_2) + P(w_3)] = [0.3,0.58]$.
Note that the set of all the answer sets of the worlds is the same as Example~\ref{ex:asp_dummy} (5 answer sets in total).
\end{example}

\begin{table}[tb]
\centering
\caption{Worlds and corresponding answer sets for Example~\ref{ex:example_pasp}. The probabilities of the worlds sum up to 1.}
\begin{tabular}{||c|c|c|c||}
  World id & Probabilistic Facts & Answer Sets & $P(w)$
  \midline
  $w_0$ & $\{\}$ & $\{\{\}\}$ & 0.42 \\
  $w_1$ & $\{a\}$ & $\{\{a,qr\}\}$ & 0.18 \\
  $w_2$ & $\{b\}$ & $\{\{b,qr\}, \{b,nqr\}\}$ & 0.28 \\
  $w_3$ & $\{a,b\}$ & $\{\{a,b,qr\}\}$ & 0.12 \\ 
\end{tabular}
\label{tab:worlds_ex_dummy}
\end{table}

The credal semantics requires that every world has at least one answer set, i.e., \red{it is satisfiable}.
If this does not hold, some of the probability mass, that is associated with the inconsistent worlds, is lost, since it is not considered neither in the formula for the lower nor in the formula for the upper probability (Equation~\eqref{eq:lower_upper_prob}).
Let us denote with $P(\mathit{inc})$ the probability of the inconsistent worlds, computed as 
\begin{equation}
\label{eq:p_inconsistent}
P(\mathit{inc}) = \sum_{w_i \ s.t. \ |AS(w_i)| = 0} P(w_i).
\end{equation}

We consider an approach also adopted in smProbLog~\citep{totis_de_raedt_kimmig_2023}: the probability of the inconsistent worlds, $P(\mathit{inc})$, is treated as a third probability value, in addition to the lower and upper probability.
In this way, $\lowerprob(q) = 1 - \upperprob(\mathit{not} \ q) - P(\mathit{inc})$ and $\upperprob(q) = 1 - \lowerprob(\mathit{not} \ q) - P(\mathit{inc})$.
In this case, if all the worlds are inconsistent, both the lower and upper probability are 0.
Let us clarify this with an example:
\begin{example}
\label{ex:inconsistent}
Consider Example~\ref{ex:example_pasp} with an additional constraint:
\begin{lstlisting}
0.3::a.
0.4::b.
qr :- a.
qr ; nqr :- b.
:- a, b.
\end{lstlisting}
Here, the constraint makes the world $w_3$ of Table~\ref{tab:worlds_ex_dummy} inconsistent as the answer set $\{a,b,qr\}$ violates the constraints.
Thus $P(\mathit{inc}) = P(w_3) = 0.12$.
Consider the query $qr$.
We have $\lowerprob(qr) = P(w_1) = 0.18$, $\upperprob(qr) = P(w_1) + P(w_2) = 0.18 + 0.28 = 0.46$, $\lowerprob(\mathit{not} \ qr) = P(w_0) = 0.42$, and $\upperprob(\mathit{not} \ qr) = P(w_0) + P(w_2) = 0.7$.
Now note that $\lowerprob(qr) = 1 - \upperprob(\mathit{not} \ qr) - P(\mathit{inc}) = 1 - 0.7 - 0.12 = 0.18$.
\end{example}

\section{DTProbLog}
\label{sec:dtproblog}
DTProbLog extends the ProbLog language with a set $D$ of (possibly non-ground) decision atoms represented with the syntax $?::d$ where $d$ is an atom, and a set $U$ of utility attributes of the form $u \to r$, where $r \in \mathbb{R}$ is the reward obtained when the utility atom $u$ is satisfied.
In the rest of the paper, we will use the terms utility and reward interchangeably and we will use the notation $utility(u,r)$ in code snippets to denote utility attributes.
A set of decision atoms defines a \textit{strategy} $\sigma$.
There are $2^{|D|}$ possible strategies.
A DTProbLog program is a tuple $(\mathcal{P},U,D)$ where $\mathcal{P}$ is a ProbLog program, $U$ is a set of utility attributes, and $D$ a set of decision atoms.
Given a strategy $\sigma$, adding all decision atoms from $\sigma$ to $\mathcal{P}$ yields a ProbLog program $\mathcal{P}_\sigma$.
The utility of a strategy $\sigma$, $\Util(\sigma)$, is given by:
\begin{equation}
\label{eq:strategy_dtproblog}
\Util(\sigma) = \sum_{w_{\sigma} \in P_{\sigma}} P(w_{\sigma}) \cdot R(w_{\sigma}) 
\end{equation}
where
\begin{equation}
R(w_{\sigma}) = \sum_{(u \to r) \in U, \ w_{\sigma} \models u} r.
\end{equation}
That is, the utility of a strategy $\sigma$ is the sum of the probabilities of the worlds $w_\sigma$ of the ProbLog program $P_\sigma$ identified by $\sigma$ multiplied by the reward of each world.
The reward of a world $w$, $R(w)$, is computed as the sum of the rewards of the utility attributes true in $w$.
Equivalently, the task can also be expressed as:
\begin{equation}
\Util(\sigma) = \sum_{(u \to r) \in U} r \cdot P_{\sigma}(u) = \sum_{(u \to r) \in U} r \cdot \sum_{w \models u} P_{\sigma}(w)
\end{equation}
where $P_\sigma(w)$ is computed with Equation~\eqref{eq:world_probability} by considering the ProbLog program $\mathcal{P}_\sigma$.
The goal of the decision theory task is to find the strategy that maximizes the utility, i.e.,
\begin{equation}
\label{eq:optimal_strategy_dtproblog}
\sigma^* = \argmax_{\sigma} \Util(\sigma).
\end{equation}
To clarify, consider Example~\ref{ex:problog_program}:
\begin{example}
\label{ex:problog_program}
The following DTProbLog program has two probabilistic facts, two decision atoms, and three utility attributes.
\begin{lstlisting}
0.1::a.
0.7::b.
? :: da.
? :: db.
q :- da, a.
q :- db, b.
utility(q,4).
utility(da,-3).
utility(db,-2).
\end{lstlisting}
There are $2^2 = 4$ possible strategies that we indicate with $\sigma_{\emptyset}$, $\sigma_{da}$, $\sigma_{db}$, and $\sigma_{dadb}$, each one defining a different ProbLog program.
With $\sigma_{\emptyset} = \{\}$, both decision atoms are not selected and we get a utility of 0.
With $\sigma_{da} = \{da\}$, we get $P(q) = 0.1$, $P(da) = 1$, so $\Util(\sigma_{da}) = 0.1 \cdot 4 - 1 \cdot 3 = -2.6$.
With $\sigma_{db} = \{db\}$, we get $P(q) = 0.7$, $P(db) = 1$, so $\Util(\sigma_{db}) = 0.7 \cdot 4 - 1 \cdot 2 = 0.8$.
With $\sigma_{dadb} = \{da,db\}$, we get $P(q) = 0.73$, $P(da) = 1$, $P(db) = 1$, so $\Util(\sigma_{dadb}) = 0.73 \cdot 4 - 1 \cdot 2 - 1 \cdot 3 = -2.08$.
Overall, the strategy that maximizes the utility is $\sigma_{db} = \{db\}$.
\end{example}

A DTProbLog program is converted into a compact form based on Algebraic Decision Diagram (ADD)~\citep{DBLP:journals/fmsd/BaharFGHMPS97} with a process called knowledge compilation~\citep{DBLP:journals/jair/DarwicheM02}, extensively adopted in Probabilistic Logic Programming~\citep{DBLP:conf/ijcai/RaedtKT07,Rig23-BK}.
ADDs are an extension of Binary Decision Diagrams (BDDs)~\citep{akers1978binary}.
BDDs are rooted directed acyclic graphs where there are only two terminal nodes, 0 and 1.
Every internal node (called decision node) is associated with a Boolean variable and has two children, one associated with the assignment true to the variable represented by the node and one associated with false.
Several additional imposed properties such as variable ordering allow one to compactly represent large search spaces, even if finding the optimal ordering of the variables that minimize the size of the BDD is a hard task~\citep{meinel1994complexity}.
In ADDs, leaf nodes may be associated with elements belonging to a set of constants (for example natural numbers) instead of only 0 and 1, that has been proved effective in multiple 
scenarios~\citep{DBLP:journals/fmsd/BaharFGHMPS97}.

\section{Second Level Algebraic Model Counting}
\label{sec:2amc}
\cite{DBLP:journals/tplp/KieselTK22} introduced Second Level Algebraic Model Counting (2AMC), needed to solve \red{tasks such as} MAP inference~\citep{DBLP:conf/ilp/ShterionovRVKMJ14} and Decision theoretic inference~\citep{DBLP:conf/aaai/BroeckTOR10}, in Probabilistic Logic Programming, and inference in smProbLog~\citep{totis_de_raedt_kimmig_2023} programs.
These problems are characterized by the need for two levels of Algebraic Model Counting (2AMC)~\citep{10.1016/j.jal.2016.11.031}.

The ingredients of a 2AMC problem are:
\begin{itemize}
  \item a propositional theory $\Pi$;
  \item a partition $(V_i,V_o)$ of the variables in $\Pi$;
  \item two commutative semirings~\citep{10.5555/1386688} $\mathcal{R}_{i} = (R^{i},\oplus^i,\otimes^i,n_{\oplus}^i,n_{\otimes}^i)$ and $\mathcal{R}_{o} = (R^{o},\oplus^o,\otimes^o,n_{\oplus}^o,n_{\otimes}^o)$;
  \item two weight functions, $w_{i}$ and $w_{o}$, associating each literal of the program with a weight; and
  \item a transformation function $f$ mapping the values of $R^i$ to those of $R^o$
\end{itemize}
Let us denote with $T$ the tuple $(\Pi,V_i,V_o,\mathcal{R}_i,\mathcal{R}_o,w_i,w_o,f)$.
The task requires solving:
\begin{align}
  \label{eq:2amc}
  \begin{split}  
    2AMC(T) =& 
    \bigoplus\nolimits_{I_{o} \in \mu(V_{o})}^{o} 
    \bigotimes\nolimits^{o}_{a \in I_{o}} 
    w_{o}(a) 
    \otimes^{o}
    f(
      \bigoplus\nolimits_{I_{i} \in \varphi(\Pi \mid I_{o})}^{i} \bigotimes\nolimits^{i}_{b \in I_{i}} w_{i}(b)  
    )  
  \end{split}
\end{align}
where $\mu(V_{o})$ is the set of possible assignments to the variables in $V_{o}$ and $\varphi(\Pi \mid I_{o})$ is the set of possible assignments to the variables in $\Pi$ that satisfy $I_{o}$.
In practice, for every possible assignment of the variables $V_o$, we need to solve a first AMC task on the variables $V_i$.
Then, the transformation function maps the obtained values into elements of the outer semiring and we need to solve a second AMC task, this time by considering $V_o$.

Within 2AMC, the DTProbLog task can be solved by considering~\citep{DBLP:journals/tplp/KieselTK22} 
\begin{itemize}
  \item as $V_o$ the decision atoms and as $V_i$ the remaining literals;
  \item as inner semiring the gradient semiring~\citep{DBLP:conf/acl/Eisner02} $\mathcal{R}_i = (\mathbb{R}^2, +, \otimes, (0,0), (0,1))$
  where $+$ is component-wise and $(a_0,b_0) \otimes (a_1,b_1) = (a_0 \cdot a_1, a_0 \cdot b_1 + a_1 \cdot b_0)$;
  \item as inner weight function $w_i$ mapping a literal $v$ to $(p,0)$ if $v = a$ where $a$ is a probabilistic fact $p::a$, to $(1-p,0)$ if $v = not \ a$ where $a$ is a probabilistic fact $p::a$, and all the other literals to $(1,r)$ where $r$ is their utility;
  \item as transformation function $f(p,u) = (u,\{\})$ if $p \neq 0$ and $f(0,u) = (-\infty,D)$;
  \item as outer semiring $\mathcal{R}_o = (\mathbb{R} \times 2^{|D|}, \oplus, \otimes, (-\infty,D), (0,\{\}))$ where $(a_0,b_0) \oplus (a_1,b_1)$ is equal to $(a_0,b_0)$ if $a_0 > a_1$, otherwise $(a_1,b_1)$ and $(a_0,b_0) \otimes (a_1,b_1) = (a_0 + a_1, b_0 \cup b_1)$; and
  \item as outer weight function $w_o = (0,\{a\})$ if $a$ is a decision atom, $(0,\{\})$ otherwise.
\end{itemize}

\cite{DBLP:journals/tplp/KieselTK22} also extended the aspmc tool~\citep{DBLP:conf/kr/EiterHK21} to solve 2AMC tasks.
aspmc converts a program into a tractable circuit (knowledge compilation) by first grounding it and then by generating a propositional formula such that the answer sets of the original program are in one-to-one correspondence with the models of the formula.
The pipeline is the following: first, aspmc breaks cycles in the program using Tp-unfolding~\citep{DBLP:conf/kr/EiterHK21}, which draws inspiration from Tp-Compilation~\citep{tp-compilaton}.
The obtained acyclic program is then translated to a propositional formula by applying a treewidth-aware version of Clark's Completion similar to that of~\cite{HECHER2022103651}.
Lastly, it solves the algebraic answer set counting task by leveraging knowledge compilers such as c2d~\citep{DBLP:conf/ecai/Darwiche04}.
The result of the compilation is a circuit in negation normal form (NNF).
An NNF is a rooted directed acyclic graph where leaf nodes are associated with a literal or a truth value (true or false) and internal nodes are associated with conjunctions or disjunctions.
Usually, sd-DNNF are considered, i.e., NNF with three additional properties.
If we denote with $V(n)$ all the variables that appear in the subgraph with root $n$, these are 
i) $V(n_a) \cap V(n_b) = \emptyset$ for each children $n_a$ and $n_b$ of an internal node associated with a conjunction (decomposability property);
ii) $n_a \land n_b$ is inconsistent for each children $n_a$ and $n_b$ of an internal node associated with a disjunction (determinism property); and 
iii) $V(n_a) = V(n_b)$ for each child $n_a$ and $n_b$ of an internal node associated with a disjunction (smoothness property).

To handle 2AMC, aspmc has been extended to perform Constrained Knowledge Compilation~\citep{DBLP:conf/ijcai/OztokD15}.
The idea is to compile the underlying logical part of the program into a tractable circuit representation over which it can evaluate the given 2AMC instance in polynomial time in its size.
Naturally, since 2AMC is harder than (weighted) model counting, one cannot use any sd-DNNF here. Instead, aspmc uses X-first circuits which constrain the order in which variables are decided.
Namely, the outer variables need to be decided first, before the inner variables can be decided.

Such ordering constraints can severely limit our options during compilation and, thus, often lead to much larger circuits.
To alleviate this,~\cite{DBLP:journals/tplp/KieselTK22} introduced X/D-first circuits that allow for less strict variable orders by using the \textit{definability} property (Definition~\ref{def:definability}).

\begin{definition}[Definability~\citep{10.5555/3060621.3060726}]
\label{def:definability}
A variable $a$ is \emph{defined} by a set of variables $X$ with respect to a theory $T$ if for every assignment $x$ of $X$ it holds that $x \cup T \models a$ or $x \cup T \models not \ a$. 
We denote the set of variables that are not in $X$ and defined by $X$ with respect to $T$ by $D(T, X)$.
\end{definition}
\begin{example}
In the program
\begin{lstlisting}
a :- b, not na.
c :- a, b.
na :- b, not a. 
0.5::b. 
\end{lstlisting}
The atom $c$ is defined in terms of $a,b$ and in terms of $na, b$.
Furthermore $a$ is defined by $na, b$ and $na$ is defined by $a, b$.
Only $b$ is not defined by any other variables.
\end{example}
Intuitively, a variable $y$ defined by $X$ can be seen to represent the truth value of a propositional formula over the variables in $X$.
Therefore, deciding $y$ is no different than making a complex decision over the variables in $X$.
It was shown that under weak conditions on the used semirings and transformation function, this also preserves the possibility of tractable evaluation of 2AMC instances over the resulting circuit~\citep{DBLP:journals/tplp/KieselTK22}.
During the evaluation of a 2AMC task, the variables of the circuit are split into two sets, call them $X$ and $Y$.
An internal node $n$ is termed \textit{pure} if $V(n) \subseteq X \cup D(n,X)$ or $V(n) \subseteq Y$, \textit{mixed} otherwise.
An NNF is X/D-first if, for each and node $n$, all its children are pure or one child $n_i$ is mixed, all the others $n_j$ are pure and $V(n_j) \subseteq X \cup D(n_i,X)$.
This additional property allows handling the variables of $X$ and $Y$ separately but allows them to be decided somewhat intertwined, whenever variables in $Y$ are defined by the variables in $X$.
\cite{korhonen2021integrating} have shown that it is highly beneficial to determine the order in which variables are decided from a so called tree decomposition of the logical theory.
Here, the decomposition intuitively provides guidance on how the problem can be split into smaller sub problems.
The obtained circuit is evaluated bottom up.

\section{Representing Decision Theory Problems with Probabilistic Answer Set Programming}
\label{sec:dt_pasp}
Following DTProbLog~\citep{DBLP:conf/aaai/BroeckTOR10}, we use $\utility(a,r)$ to denote utility attributes where $a$ is an atom and $r \in \mathbb{R}$ indicates the utility of satisfying it.
For example, with $\utility(a,-3.3)$ we state that if $a$ is satisfied we get a utility of -3.3.
A negative utility represents, for example, a cost, while a positive utility represents a gain.
We use the functor $\decision$ to denote decision atoms.
For example, with $\decision a$ we state that $a$ is a decision atom.
A decision atom indicates that we can choose whether to perform or not to perform the specified action.
We consider only ground decision atoms.
\begin{definition}
\label{def:DTPASP}
A \textit{decision theory probabilistic answer set program} DTPASP is a tuple $(\mathcal{P},D,U)$ where $\mathcal{P}$ is a probabilistic answer set program, $D$ is the set of decision atoms, and $U$ is the set of utility attributes.
\end{definition}

Since in PASP queries are associated with a lower and an upper probability, in DTPASP we need to consider lower and upper rewards and look for the strategies that maximize them.
A strategy is, as in DTProbLog, a subset of the possible actions. Having fixed a strategy $\sigma$, we obtain a PASP $P_\sigma$ that generates a set of worlds.
Each answer set $A$ of a world $w_{\sigma}$ of $P_{\sigma}$ is associated with a reward given by the sum of the utilities of the atoms true in it:
\begin{equation}
\label{eq:answer_set_reward}
R(A) = \sum_{(a,r) \in U,\ a \in A} r.
\end{equation}
Let us call $\underline{R}(w_\sigma)$ the minimum of the rewards of an answer set of $w_\sigma$ and $\overline{R}(w_\sigma)$ the maximum of the rewards of an answer set of $w_\sigma$.
That is:
\begin{equation}
\label{eq:min_max_reward}
\begin{aligned}
  \underline{R}(w_\sigma) &= \min_{A \in AS(w_{\sigma})} R(A), \\
  \overline{R}(w_\sigma) &= \max_{A \in AS(w_{\sigma})} R(A).
\end{aligned}
\end{equation}
Since we impose no constraints on how the probability mass of a world is distributed among its answer sets, we can obtain the minimum utility from a world by assigning all the mass to the answer set with the minimum reward and the maximum utility from a world by assigning all the mass to the answer set with the maximum reward. 
The first is the worst case scenario and the latter is the best case scenario.
The expected minimum $\underline{U}(w_\sigma)$ and maximum $\overline{U}(w_\sigma)$ utility from $w_\sigma$ are thus:
\begin{equation}
\label{eq:min_max_utility}
\begin{aligned}
  \underline{U}(w_\sigma) = \underline{R}(w_\sigma)\cdot P(w_\sigma), \\
  \overline{U}(w_\sigma) = \overline{R}(w_\sigma)\cdot P(w_\sigma).
\end{aligned}
\end{equation}
A strategy $\sigma$ is associated with a lower $\lowerutil(\sigma)$ and upper $\upperutil(\sigma)$ utility given by, respectively:
\begin{equation}
\label{eq:utility_optimization}
\begin{aligned}
  \lowerutil(\sigma) &= \sum_{w_\sigma} \underline{U}(w_\sigma), \\
  \upperutil(\sigma) &= \sum_{w_\sigma} \overline{U}(w_\sigma).
\end{aligned}
\end{equation}
Let us indicate the range of expected utilities with $\Util(\sigma) = [\lowerutil(\sigma),\upperutil(\sigma)]$.
Finally, the goal of decision theory in PASP is to find the two strategies that maximize the lower and upper bound of the expected utility, let us call them the lower and upper strategies, respectively, i.e.:
\begin{equation}
\label{eq:solution_dtpasp_task}
\begin{aligned}
\underline{\sigma}^* &= \argmax_{\sigma} \lowerutil(\sigma), \\
\overline{\sigma}^* &= \argmax_{\sigma} \upperutil(\sigma).
\end{aligned}
\end{equation}

Furthermore, if every world in a PASP has exactly one answer set:
i) $\lowerutil(\sigma)$ and $\upperutil(\sigma)$ of Equation~\eqref{eq:utility_optimization} coincide and
ii) Equation~\eqref{eq:strategy_dtproblog} and Equation~\eqref{eq:utility_optimization} return the same value.
In fact, if every world has exactly one answer set, the reward for each world is the same if we consider the best or worst case scenario, thus the sum of the rewards coincides in the two cases.
Similar considerations hold for ii).

\subsection{Examples}
\label{subsec:examples}

In this section, we will discuss a practical application of DTPASP to the viral marketing scenario. 
We consider here the problem of computing the upper strategy, but considerations for the lower strategy are analogous.
First, let us introduce a running example that will be discussed multiple times across the paper.
\begin{example}[Running Example.]
\label{ex:running_dummy}
Consider a variation of Example~\ref{ex:example_pasp} with 2 probabilistic facts and 2 decision atoms.
\begin{lstlisting}
0.3::a. 0.4::b.
decision da. decision db.
utility(qr,2). utility(nqr,-12).
qr:- da, a. 
qr;nqr:- db, b.
\end{lstlisting}
There are 4 possible strategies that we indicate with $\sigma_{\emptyset,da,db,dadb}$.
With $\sigma_{\emptyset} = \{\}$ we have the PASP
\begin{lstlisting}
0.3::a. 0.4::b. 
qr:- da, a. 
qr;nqr:- db, b.
\end{lstlisting}
This program has 4 worlds, \red{listed in Table~\ref{tab:ex_running_dtpasp_sigma_empty}}, each having no answer sets where the utility atoms are true, so $\Util(\sigma_{\emptyset}) = [0,0]$.

With $\sigma_{da}$ we have the PASP
\begin{lstlisting}
0.3::a. 0.4::b. 
da.
qr:- da, a. 
qr;nqr:- db, b.
\end{lstlisting}
\red{The worlds together with rewards and answer sets are listed in Table~\ref{tab:ex_running_dtpasp_sigma_da}}.
This strategy has a utility of $\Util(\sigma_{da}) = [0.36 + 0.24, 0.36 + 0.24] = [0.6,0.6]$.

With $\sigma_{db}$ we have the PASP
\begin{lstlisting}
0.3::a. 0.4::b. 
db.
qr:- da, a. 
qr;nqr:- db, b.
\end{lstlisting}
\red{The worlds together with rewards and answer sets are listed in Table~\ref{tab:ex_running_dtpasp_sigma_db}}.
This strategy has a utility of $\Util(\sigma_{db}) = [-3.36 - 1.44, 0.56 + 0.24] = [-4.8,0.8]$.

With $\sigma_{dadb}$ we have the PASP
\begin{lstlisting}
0.3::a. 0.4::b. 
da. db.
qr:- da, a. 
qr;nqr:- db, b.
\end{lstlisting}
\red{The worlds together with rewards and answer sets are listed in Table~\ref{tab:ex_running_dtpasp_sigma_dadb}}.
This strategy has a utility of $\Util(\sigma_{dadb}) = [0.36-3.36+0.24, 0.36+0.56+0.24] = [-2.76,1.16]$.
Overall, the strategy $\sigma_{da}$ yields the highest lower bound (0.6) for the utility while $\sigma_{dadb}$ yields the highest upper bound (1.16) for the utility.
That is, in the worst case we obtain a reward of 0.6 and in the best case a reward of 1.16.
\end{example}

\begin{table}[t]
  \centering
  \caption{\red{Answer sets, worlds, and rewards for the strategy $\sigma_{\emptyset} = \{\}$ of Example~\ref{ex:running_dummy}.}}
  \red{
  \begin{tabular}{||c|c|c|c|c||}
    World id & Probabilistic Facts & Answer Sets & $P(w)$ & $R(w)$
    \midline
    $w_0^{\sigma_{\emptyset}}$ & $\{\}$     & $\{\{\}\}$ & 0.42 & [0,0] \\
    $w_1^{\sigma_{\emptyset}}$ & $\{a\}$    & $\{\{\}\}$ & 0.18 & [0,0] \\
    $w_2^{\sigma_{\emptyset}}$ & $\{b\}$    & $\{\{\}\}$ & 0.28 & [0,0] \\
    $w_3^{\sigma_{\emptyset}}$ & $\{a,b\}$  & $\{\{\}\}$ & 0.12 & [0,0] \\ 
  \end{tabular}
  }
  \label{tab:ex_running_dtpasp_sigma_empty}
\end{table}

\begin{table}[t]
  \centering
  \caption{\red{Answer sets, worlds, and rewards for the strategy $\sigma_{\sigma_{da}} = \{da\}$ of Example~\ref{ex:running_dummy}.}}
  \red{
  \begin{tabular}{||c|c|c|c|c||}
    World id & Probabilistic Facts & Answer Sets & $P(w)$ & $R(w)$
    \midline
    $w_0^{\sigma_{da}}$ & $\{\}$    & $\{\{da\}\}$ & 0.42 & [0,0]\\
    $w_1^{\sigma_{da}}$ & $\{a\}$   & $\{\{a,da,qr\}\}$ & 0.18 & [0.36,0.36] \\
    $w_2^{\sigma_{da}}$ & $\{b\}$   & $\{\{b,da\}\}$ & 0.28 & [0,0] \\
    $w_3^{\sigma_{da}}$ & $\{a,b\}$ & $\{\{a,b,da,qr\}\}$ & 0.12 & [0.24,0.24] \\ 
  \end{tabular}
  }
  \label{tab:ex_running_dtpasp_sigma_da}
\end{table}

\begin{table}[t]
  \centering
  \caption{\red{Answer sets, worlds, and rewards for the strategy $\sigma_{\sigma_{db}} = \{db\}$ of Example~\ref{ex:running_dummy}.}}
  \red{
  \begin{tabular}{||c|c|c|c|c||}
    World id & Probabilistic Facts & Answer Sets & $P(w)$ & $R(w)$
    \midline
    $w_0^{\sigma_{db}}$ & $\{\}$ & $\{\{db\}\}$ & 0.42 & [0,0] \\
    $w_1^{\sigma_{db}}$ & $\{a\}$ & $\{\{a,db\}\}$ & 0.18 & [0,0] \\
    $w_2^{\sigma_{db}}$ & $\{b\}$ & $\{\{b,db,qr\},\{b,db,nqr\}\}$ & 0.28 & [-3.36,0.56] \\
    $w_3^{\sigma_{db}}$ & $\{a,b\}$ & $\{\{a,b,db,qr\},\{a,b,db,nqr\}\}$ & 0.12 & [-1.44,0.24] \\ 
  \end{tabular}
  }
  \label{tab:ex_running_dtpasp_sigma_db}
\end{table}

\begin{table}[t]
  \centering
  \caption{\red{Answer sets, worlds, and rewards for the strategy $\sigma_{\sigma_{dadb}} = \{da,db\}$ of Example~\ref{ex:running_dummy}.}}
  \red{
  \begin{tabular}{||c|c|c|c|c||}
    World id & Probabilistic Facts & Answer Sets & $P(w)$ & $R(w)$
    \midline
    $w_0^{\sigma_{dadb}}$ & $\{\}$ & $\{\{da,db\}\}$ & 0.42 & [0,0] \\
    $w_1^{\sigma_{dadb}}$ & $\{a\}$ & $\{\{a,qr,da,db\}\}$ & 0.18 & [0.36,0.36] \\
    $w_2^{\sigma_{dadb}}$ & $\{b\}$ & $\{\{b,da,db,qr\},\{b,da,db,nqr\}\}$ & 0.28 & [-3.36,0.56] \\
    $w_3^{\sigma_{dadb}}$ & $\{a,b\}$ & $\{\{a,b,qr,da,db\}\}$ & 0.12 & [0.24,0.24] \\ 
  \end{tabular}
  }
  \label{tab:ex_running_dtpasp_sigma_dadb}
\end{table}

\begin{example}
Consider a marketing scenario, where people go shopping with a given probability (probabilistic facts).
We need to decide which people to target with a marketing action (decision atoms).
If these people go shopping and are targeted with a personalized advertisement, then they can buy some products.
These products have an associated utility, as the target operation does (because, for example, targeting involves a cost).
Moreover, suppose we have a constraint \red{imposing a limit on the quantity of a certain product.
For example, suppose we want lo limit the sales of spaghetti to one unit, because the company has liimited stock of that product.}
If we consider 2 people, Anna and Bob, the just described scenario can be represented with the following program.
\label{ex:motivating}
\begin{lstlisting}
0.8::shops(anna). 0.5::shops(bob).

decision target(anna). decision target(bob).

buy(spaghetti,anna) ; buy(steak,anna) :- 
    shops(anna), target(anna).
buy(spaghetti,bob) ; buy(beans,bob) :- 
    shops(bob), target(bob).

utility(target(anna),-2). utility(target(bob),-2).
utility(buy(spaghetti,anna),6). utility(buy(steak,anna),1).
utility(buy(spaghetti,bob),7). utility(buy(beans,bob),7).

(*@\red{:- \#count\{X : buy(spaghetti,X)\} > 1.} @*)
\end{lstlisting}

By applying the same approach of Example~\ref{ex:motivating}, we have 4 possible strategies, $\sigma_{00} = \{\}$,
$\sigma_{01} = \{\mathit{target(anna)}\}$,
$\sigma_{10} = \{\mathit{target(bob)}\}$, and
$\sigma_{11} = \{\mathit{target(anna)}, \mathit{target(bob)}\}$
with
\red{
$\Util(\sigma_{00}) = [0,0]$, 
$\Util(\sigma_{01}) = [-1.2,2.8]$, 
$\Util(\sigma_{10}) = [1.5,1.5]$, and 
$\Util(\sigma_{11}) = [0.3,4.3]$.
Thus, $\sigma_{10} = \{\mathit{target(bob)}\}$ is the strategy that maximizes the lower bound of the utility while the strategy $\sigma_{11} = \{\mathit{target(anna)}, \mathit{target(bob)}\}$ is the strategy that maximizes the upper bound of the utility.
So, in the worst case, by targeting Bob, we get a reward of 1.5.
Similarly, in the best case, if we target both Anna and Bob we get a reward of 4.3.}
\end{example}

In Example~\ref{ex:running_dummy}, for every strategy, every world is satisfiable \red{(it has at least one answer set)}.
However, this may not always be the case.
Consider these three different scenarios:
i) a constraint is such that for some strategies, every world has no answer sets, while for all the remaining strategies every world has at least one answer set, i.e., \red{the constraint} involves only decision atoms; 
ii) a constraint is such that some of the worlds in some strategies have no answer sets, i.e., \red{the constraint} involves decision atoms and probabilistic facts; 
iii) a constraint is such that for all the strategies the same worlds have no answer sets, i.e., \red{the constraint} involves only probabilistic facts.
These \red{possible sources of inconsistencies} are clarified with the following three examples.

\begin{example}
\label{ex:running_dummy_constr_da_db}
Consider Example~\ref{ex:running_dummy} with the additional constraint $\impl da, db$ that prevents $da$ and $db$ to be performed simultaneously.
Here, all the probabilistic answer set programs for all the strategies have a credal semantics, except for the one with both $da$ and $db$ true ($\sigma_{dadb}$), which leads to inconsistent worlds.
\end{example}

\begin{example}
\label{ex:running_dummy_constr_db_a}
Consider Example~\ref{ex:running_dummy} with the additional constraint $\impl db, a$ that prevents $db$ and $a$ to be true simultaneously.
Consider $\sigma_{db}$.
Only two worlds are satisfiable, namely $w_0 = \{\}$ and $w_2 = \{b\}$, so all the worlds of the PASP identified by the strategy $\sigma_{db}$ are inconsistent.
Therefore, we have to decide whether to discard this strategy or to keep it and consider the inconsistent worlds in some way.
Similar considerations can be applied to $\sigma_{dadb}$.
In this example, some strategies yield a consistent PASP, others don't.
\end{example}

\begin{example}
\label{ex:running_dummy_constr_a_b}
Consider Example~\ref{ex:running_dummy} with the additional constraint $\impl a, b$ that prevents $a$ and $b$ to be true simultaneously.
Differently from Example~\ref{ex:running_dummy_constr_db_a}, here every strategy $\sigma_{ij}, \ i,j \in \{da,db\}$, results in a probabilistic answer set program has an inconsistent world, since world $w_3$ (Table~\ref{tab:worlds_ex_dummy}), where both $a$ and $b$ are true, is inconsistent.
Here, all strategies yield an inconsistent probabilistic answer set program.
\end{example}

The previous three examples illustrate some of the possible scenarios that may arise.
In Example~\ref{ex:running_dummy_constr_da_db}, we can discard the inconsistent probabilistic answer set program where none of the worlds are satisfiable and pick the best strategy among $\sigma_{\emptyset}$, $\sigma_{da}$, and $\sigma_{db}$.
For Examples~\ref{ex:running_dummy_constr_db_a} and~\ref{ex:running_dummy_constr_a_b}, where only some of the worlds are inconsistent, we can ignore these and proceed in analysing the remaining.

\section{Algorithms for Computing the Best Strategy in a DTPASP}
\label{sec:algorithm}
To solve the optimization problems represented in Equation~\eqref{eq:solution_dtpasp_task}, we discuss two different algorithms.
We have three different layers of complexity: 
i) the computation of the possible strategies, 
ii) the computation of the worlds, and 
iii) the computation of the answer sets with the highest and lowest reward (Equation~\eqref{eq:min_max_reward}).
Due to this, the task cannot be solved with standard optimization constructs, such as $\#\mathit{maximize}$, available in ASP solvers.

We developed a first exact algorithm that iteratively enumerates all the strategies and, for every strategy, computes the answer sets for every world.
Given a decision theoretic probabilistic answer set program with $n$ probabilistic facts and $d$ decision atoms, if every world is satisfiable, we need to generate at least $2^d \cdot 2^n = 2^{d+n}$ answer sets.
Clearly, this algorithm is feasible only for trivial domains, so we consider it only as a baseline.

We propose a second and more interesting approach.
First, the task of Equation~\eqref{eq:solution_dtpasp_task} cannot be represented as a 2AMC problem since it requires three layers of AMC (3AMC).
Thus, by extending Equation~\eqref{eq:2amc}, we define 3AMC as:
\begin{align}
  \begin{split}  
    3AMC(T) =& 
    \bigoplus\nolimits_{I_{o} \in \mu(V_{o})}^{o}
    \bigotimes\nolimits^{o}_{a \in I_{o}} 
    w_{o}(a) 
    \otimes^{o} \\ 
    & f_{mo}(
        \bigoplus\nolimits_{I_{m} \in \varphi(\Pi \mid I_{o})}^{m} \bigotimes\nolimits^{m}_{b \in I_{m}} w_{m}(b)
        \otimes^{m} \\ 
        & 
        f_{im}(
          \bigoplus\nolimits_{I_{i} \in \varphi(\Pi \mid I_{m})}^{i} \bigotimes\nolimits^{i}_{c \in I_{i}} w_{i}(c)
        )
    ).
  \end{split}
\end{align}
That is, we add a third layer of AMC on top of 2AMC, obtaining 3AMC.
For the decision theory task in probabilistic answer set programming, in the innermost layer (call it $A_i$), we have both the strategy and the world fixed, and we compute the reward for each answer set (Equation~\eqref{eq:answer_set_reward}) and find the one that minimizes and maximizes the reward.
In the middle layer (call it $A_m$), we have a fixed strategy and we need to find the probabilities of all the worlds, multiply them by the optimal rewards obtained in the previous step, and sum all these products (Equation~\eqref{eq:utility_optimization}).
Lastly, in the outer layer (call it $A_o$) we need to compute all the strategies and find the two that maximize the lower and upper utility (Equation~\eqref{eq:solution_dtpasp_task}), respectively.
Note that with only one AMC we are able to find both.
If we call $H$, $D$, and $F$, the set of all the variables, decision atoms, and probabilistic facts, respectively, for the innermost layer $A_i$ we have:
\begin{itemize}
  \item as semiring, the minmax-plus semiring $\mathcal{R}_i = (\mathbb{R}^2, minmax, +^2, (\infty,-\infty), (0,0))$, where $minmax((a_0,b_0),(a_1,b_1))$ returns the pair ($a_0$ if $a_0 < a_1$ else $a_1$, $b_0$ if $b_0 > b_1$ else $b_1$) and $+^2((a_0,b_0),(a_1,b_1)) = (a_0+a_1,b_0+b_1)$ 
  \item as variables $H \setminus D \setminus F$
  \item as weight function 
    \begin{equation*}
        w_0(a) =
        \begin{dcases*}
          (r,r) & if $(a,r) \in U,$ \\
          (0,0) & otherwise.
        \end{dcases*}
    \end{equation*}
\end{itemize}
As transformation function $f_{im}$ that maps the values of $A_i$ to $A_m$ we have $f_{im} : \mathbb{R}^2 \to \mathbb{R}^3$, $f_{im}(a,b) = (1,a,b)$.
As middle layer $A_m$ we have:
\begin{itemize}
  \item as semiring, the two-gradient semiring $\mathcal{R}_m = (\mathbb{R}^3, \oplus^G, \otimes^G, (0,0,0), (1,0,0))$ with $(a_0,b_0,c_0) \oplus^G (a_1, b_1,c_1) = (a_0 + a_1, b_0 + b_1, c_0 + c_1)$ and $(a_0,b_0,c_0) \otimes^G (a_1, b_1,c_1) = (a_0 \cdot a_1, a_0 \cdot b_1 + a_1 \cdot b_0, a_0 \cdot c_1 + a_1 \cdot c_0)$ 
  \item as variables $F$
  \item as weight function 
    \begin{equation*}
        w_1(a) =
        \begin{dcases*}
          (p,0,0) & if $a = f$ where $f$ is a probabilistic fact $p::f,$ \\
          (1-p,0,0) & if $a = not \ f$ where $f$ is a probabilistic fact $p::f,$ \\
          (1,0,0) & otherwise.
        \end{dcases*}
    \end{equation*}
\end{itemize}
Here, the first component stores the probability computed so far while the remaining two store the lower and upper bounds for the utility.
As transformation function $f_{mo}$ that maps the values of $A_m$ to $A_o$, we have
$f_{mo}(a,b,c) = (b,c,\{\},\{\})$, where $b$ and $c$ are the second and third components of the triple obtained from the previous layer.
Lastly, as outer layer $A_o$ we have:
\begin{itemize}
  \item as semiring $\mathcal{R}_o = (\mathbb{R}^2 \times {2^{|D|}}^2, \mathit{max}^4, \mathit{sum}^4, (\infty,-\infty,D,D), (0,0,\{\},\{\}))$
  \item as variables $D$
  \item as weight function
    \begin{equation*}
        w_2(a) =
        \begin{dcases*}
          (0,0,\{a\},\{a\}) & if $a$ is a decision atom,\\
          (0,0,\{\},\{\}) & otherwise.
        \end{dcases*}
    \end{equation*}
\end{itemize}
where $\mathit{max}^4((v_0,v_1,S_0,S_1),(v_a,v_b,S_a,S_b)) = (v_x,v_y,S_x,S_y)$ with 
$v_x = v_0$ if $v_0 > v_a$ else $v_a$,
$v_y = v_1$ if $v_1 > v_b$ else $v_b$,
$S_x = S_0$ if $v_0 > v_a$ else $S_a$,
$S_y = S_1$ if $v_1 > v_b$ else $S_b$
and $\mathit{sum}^4((v_0,v_1,S_0,S_1),(v_a,v_b,S_a,S_b)) = (v_0 + v_a, v_1 + v_b, S_0 \cup S_a, S_1 \cup S_b)$.
The first two components store, respectively, the value of the lower and upper strategies and the last two the decision atoms yielding these values.
Note that we assume that decision atoms do not have an associated utility.
This is not a restriction since it is always possible to mimic it by adding a rule $rda \impl da$ and a utility attribute on $rda$ for a decision atom $da$.

\subsection{\red{Implementation in aspmc}}
\label{subsec:implementation_aspmc}

Before discussing the algorithm, let us introduce some concepts.
\begin{definition}
\red{
A \emph{tree decomposition}~\citep{bodlaender1988treewidth} of a graph $G$ is a pair $(T, \chi)$, where $T$ is a tree and $\chi$ is a labeling of $V(T)$ (the set of nodes of $T$) by subsets of $V(G)$ (the set of nodes of $G$) s.t.
1) for all nodes $v \in V(G)$ there is $t \in V(T)$ s.t. $v \in \chi(t)$;
2) for every edge $\{v_1, v_2\} \in V(E)$ there exists $t \in V(T)$ s.t. $v_1, v_2 \in \chi(t)$; and
3) for all nodes $v \in V(G)$ the set of nodes $\{t \in V(T) \mid v \in \chi(t)\}$ forms a (connected) subtree of $T$.
The width of $(T, \chi)$ is $\max_{t \in V'} |\chi(t)| - 1$.
The \emph{treewidth} of a graph is the minimal width of any of its tree decompositions. }
\end{definition}
\red{Intuitively, treewidth is a measure of the distance of a graph from being a tree.
Accordingly, a graph is a tree if and only if it has treewidth $1$.
The idea behind treewidth is that problems that are simple when their underlying structure is a tree, may also be simple when they are not far from trees, i.e., have low treewidth.
Practically, we can use tree decompositions witnessing the low treewidth to decompose problems into smaller subproblems in such cases.}

\red{We assume that programs have already been translated into equivalent 3AMC instances, where the propositional theory is a propositional formula in conjunctive normal form (CNF).
CNFs are sets of \emph{clauses} $C_i$, representing their conjunction, where a clause is a set of \emph{literals}, i.e., possibly negated propositional variables, which is interpreted disjunctively.}
\begin{example}
\label{ex:running_cnf}
\red{Our running example for a CNF is the formula
\begin{align*}
    \mathcal{C}_{run} = \;\;&(v_1 \vee x_1) \wedge (\neg v_1 \vee \neg x_1) \wedge (v_2 \vee \neg v_3 \vee x_1) \wedge (v_4 \vee \neg x_1) \wedge (x_1 \vee x_2 \vee x_3) \\
    &\wedge (\neg x_2 \vee y_1) \wedge (\neg x_3 \vee y_2). 
\end{align*}
Here, $v_1, v_2, x_1, \dots$ are propositional variables, and thus, $v_1, \neg v_1$ are both literals. Furthermore, $v_1 \vee x_1$ and $\neg v_1 \vee \neg x_1$ denote clauses.}
\end{example}

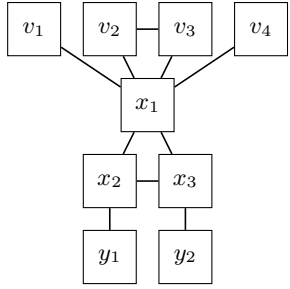
\begin{figure}[tb]
\centering
    \begin{tikzpicture}
        \node [style=node] (v1) at (0, 0) {$v_1$};
        \node [style=node] (v2) at (1, 0) {$v_2$};
        \node [style=node] (v3) at (2, 0) {$v_3$};
        \node [style=node] (v4) at (3, 0) {$v_4$};
        \node [style=node] (x1) at (1.5, -1) {$x_1$};
        \node [style=node] (x2) at (1, -2) {$x_2$};
        \node [style=node] (x3) at (2, -2) {$x_3$};
        \node [style=node] (y1) at (1, -3) {$y_1$};
        \node [style=node] (y2) at (2, -3) {$y_2$};
        \draw[-,semithick] (v1) to (x1);
        \draw[-,semithick] (v2) to (x1);
        \draw[-,semithick] (v3) to (x1);
        \draw[-,semithick] (v4) to (x1);
        \draw[-,semithick] (v2) to (v3);
        \draw[-,semithick] (x1) to (x2);
        \draw[-,semithick] (x1) to (x3);
        \draw[-,semithick] (x2) to (x3);
        \draw[-,semithick] (x2) to (y1);
        \draw[-,semithick] (x3) to (y2);
    \end{tikzpicture}
\caption{Primal graph of $\mathcal{C}_{run}$ (Example~\ref{ex:running_cnf}).}
\label{fig:primal_running}
\end{figure}

\red{For CNFs, the relevant underlying structure is often chosen as their primal graph.}
\begin{definition}
\red{Given a CNF $\mathcal{C}$, the \emph{primal graph}~\citep{oztok2014compiling} of $\mathcal{C}$ is defined as the graph $G = (V, E)$ such that $V$ is the set of propositional variables occurring in $V$ and $\{v,x\} \in E$ if $v, x$ co-occur in a clause of $\mathcal{P}$.}
\end{definition}

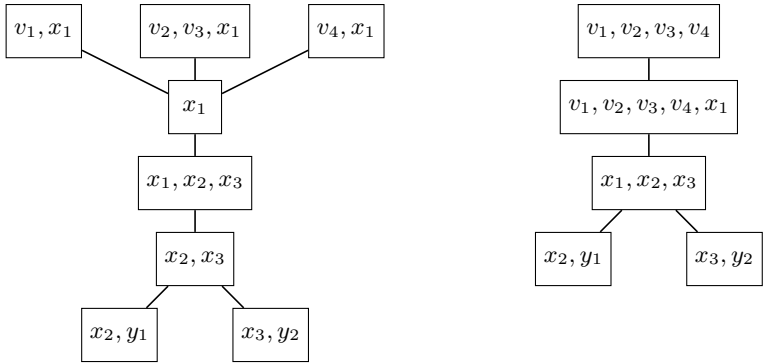
\begin{figure}[tb]
    \centering
    \begin{tikzpicture}
        \node [style=node] (t1) at (0, 0) {$v_1, x_1$};
        \node [style=node] (t2) at (2, 0) {$v_2, v_3, x_1$};
        \node [style=node] (t3) at (4, 0) {$v_4, x_1$};
        \node [style=node] (t41) at (2, -1) {$x_1$};
        \node [style=node] (t42) at (2, -2) {$x_1, x_2, x_3$};
        \node [style=node] (t43) at (2, -3) {$x_2, x_3$};
        \node [style=node] (t5) at (1, -4) {$x_2, y_1$};
        \node [style=node] (t6) at (3, -4) {$x_3, y_2$};
        
        \draw[-,semithick] (t1) to (t41);
        \draw[-,semithick] (t2) to (t41);
        \draw[-,semithick] (t3) to (t41);
        \draw[-,semithick] (t41) to (t42);
        \draw[-,semithick] (t42) to (t43);
        \draw[-,semithick] (t43) to (t5);
        \draw[-,semithick] (t43) to (t6);

        \node [style=node] (n1) at (8, 0) {$v_1, v_2, v_3, v_4$};
        \node [style=node] (n2) at (8, -1) {$v_1, v_2, v_3, v_4, x_1$};
        \node [style=node] (n3) at (8, -2) {$x_1, x_2, x_3$};
        \node [style=node] (n4) at (7, -3) {$x_2, y_1$};
        \node [style=node] (n5) at (9, -3) {$x_3, y_2$};
        
        \draw[-,semithick] (n1) to (n2);
        \draw[-,semithick] (n2) to (n3);
        \draw[-,semithick] (n3) to (n4);
        \draw[-,semithick] (n3) to (n5);
    \end{tikzpicture}
    \caption{Two tree decomposition of the graph in Figure~\ref{fig:primal_running}. Each vertex is labeled by the vertices in the corresponding bag.}
    \label{fig:primal_td_running}
\end{figure}

\begin{example}[cont.]
\red{The primal graph of $\mathcal{C}_{run}$ is given in Figure~\ref{fig:primal_running}.
Two of its tree decompositions are given in Figure~\ref{fig:primal_td_running}. 
Its treewidth is 2 (Figure~\ref{fig:primal_td_running} left). 
A smaller width is not possible here, since the graph contains a clique over three vertices ($x_1, x_2, x_3$).
The decomposition shown in Figure~\ref{fig:primal_td_running} right is not optimal as it has width 4.}
\end{example}

\red{Typically in constrained compilation, for a 3AMC instance with propositional theory $T$ and, respectively outer, middle, and inner variables $\mathbf{X}_{O},\mathbf{X}_{M},\mathbf{X}_{I}$, one would first decide all variables in $\mathbf{X}_{O}$, then all variables in $\mathbf{X}_{M}$, and finally all variables in $\mathbf{X}_{I}$.
In general, this is necessary to preserve correctness for 3AMC evaluation.
However, the idea of~\cite{DBLP:journals/tplp/KieselTK22} allows us to perform constrained compilation along a tree decomposition, where these ordering constraints are relaxed using defined variables.}

\red{Given a CNF $\mathcal{C}$ and a partition $\mathbf{X}_{O},\mathbf{X}_{M},\mathbf{X}_{I}$ of its variables $\mathbf{X}$, a tree decomposition $(T, \chi)$ of the primal graph of $\mathcal{C}$ is a $\mathbf{X}_{O} > \mathbf{X}_{M} > \mathbf{X}_{I}/D$-tree decomposition, if 
\begin{enumerate}
    \item there exists $t_{O} \in V(T)$ such that
    \begin{enumerate}
        \item $\chi(t_{O}) \subseteq \mathbf{X}_{O} \cup D(C, \mathbf{X}_{O})$, and 
        \item every path from $\mathbf{X}_{O}$ to $\mathbf{X}\setminus (\mathbf{X}_{O} \cup D(C, \mathbf{X}_{O}))$ in the primal graph of $\mathcal{C}$ uses a vertex from $\chi(t_{O})$
    \end{enumerate}
    \item and there exists $t_{M} \in V(T)$ such that 
    \begin{enumerate}
        \item $\chi(t_{M}) \subseteq \mathbf{X}_{M} \cup D(C, \mathbf{X}_{O} \cup \mathbf{X}_{M})$, and 
        \item every path from $\mathbf{X}_{M}$ to $\mathbf{X}\setminus (\mathbf{X}_{O} \cup \mathbf{X}_{M}\cup D(C, \mathbf{X}_{O} \cup \mathbf{X}_{M}))$ in the primal graph of $\mathcal{C}$ uses a vertex from $\chi(t_{M})$.
    \end{enumerate} 
\end{enumerate}}

\red{This property guarantees the following: during constrained compilation, we can decide all variables in $\chi(t_{O})$ (as in 1. (a)) and be sure that the remaining CNF decomposes into strongly connected components that either contain \emph{only} variables in $\mathbf{X}_O \cup D(C, \mathbf{X}_{O})$ or \emph{no} variables in $\mathbf{X}_O \cup D(C, \mathbf{X}_{O})$.
Intuitively, we separate the outer variables from the remaining ones (modulo definition).
An analogous statement can be made for $\chi(t_{M})$ (as in 2. (a)). }

\red{The same arguments as in~\citep{DBLP:journals/tplp/KieselTK22} allow us to conclude that we can use $\mathbf{X}_{O} > \mathbf{X}_{M} > \mathbf{X}_{I}/D$-tree decompositions to compile 3AMC instances and obtain a correct result.}

\red{As with 2AMC, definability helps in solving the task.}
\begin{example}[cont.]
\red{Consider a 3AMC instance with propositional theory $\mathcal{C}_{run}$, and resp.\ outer, middle, and inner variables 
\[\mathbf{X}_{O} = \{v_1, v_2, v_3, v_4\}, \mathbf{X}_{M} = \{x_1, x_2, x_3\}, \mathbf{X}_{I} = \{y_1, y_2\}.\]}

\red{If we disregard definability, an optimal tree decomposition that implements order constraints is the right decomposition in Figure~\ref{fig:primal_td_running}.
We have as $t_{O}$ the first bag from the top and as $t_{M}$ the third bag from the top.
This is optimal, since the minimal set of variables from $\mathbf{X}_{O}$ to separate $\mathbf{X}_{O}$ from $\mathbf{X}_{M} \cup \mathbf{X}_{I}$ is $\mathbf{X}_{O}$.
Additionally, since we need to satisfy all properties of a tree decomposition, we need a bag that contains all variables in $\mathbf{X}_{O}$ and $x_1$.}

\red{On the other hand, if we exploit definability, the left decomposition in Figure~\ref{fig:primal_td_running} is allowed.
Due to the clauses $v_1 \vee x_1$, and $\neg v_1 \vee \neg x_1$, $x_1$ is defined in terms of $v_1$ and therefore $x_1 \in D(\mathcal{C}_{run}, \mathbf{X}_{O})$.
Then, we can choose the second bag from the top as $t_{O}$, and choose the third bag from the top as $t_{M}$.
Clearly, $\{x_1\}$ already separates $\mathbf{X}_{O}$ from the non-defined variables of the variables.}
\end{example}

\red{To generate $\mathbf{X}_{O} > \mathbf{X}_{M} > \mathbf{X}_{I}/D$-tree decompositions, we proceed by using Algorithm~\ref{alg:td_from_cnf} on the CNF generated by aspmc.
We first choose the variables to decide first in order to separate the outer variables $\outX$ from the remaining ones (lines 1--3), and generate a decomposition $TD_O$ for $G_O$, the part of the primal graph $G$ that we ``cut off'' using the separator $S_O$ (lines 4--7).
Here, $\textsc{SCC}(G)$ denotes the set of strongly connected components of $G$, and $\textsc{Clique}(V)$ the complete graph over the vertices $V$.
Furthermore, $\textsc{MinimumSeparator}(G,V,W)$ is a polynomial subroutine based on a standard min-cut/max-flow algorithm that computes a minimum set of vertices from $V \cup W$ that separates all vertices in $W$ from any other vertices in $G$, and $\textsc{TreeDecomposition}(G)$ computes a decomposition from $G$.\footnote{Practically, we use flow-cutter~\citep{hamann2018graph} to compute decompositions.}
We repeat this idea for the inner variables (lines 8--13) obtaining a decomposition $TD_M$. 
Note that in this step we consider variables that are defined in terms of both the outer and the middle variables, allowing additional freedom.
Next, we generate a decomposition $TD_I$ for the remaining inner variables, and finally combine all three decompositions to one final decomposition of $C$.
For this step it is important that we add the cliques to $G_O, G_M$, and $G_I$, since these intuitively mark the points at which we connect the partial decompositions.}

\begin{algorithm}[t]
\makeatletter
\makeatother
\textbf{Input} A CNF $C$ and a partition $\outX, \midX, \innX$ of its variables.\\
\textbf{Output} A tree decomposition enforcing $\outX > \midX >\innX/D$-ordering constraints.
\begin{algorithmic}[1]
\State $G$ = \textsc{PrimalGraph}($C$)
\State $S_{O}$ = \textsc{MinimumSeparator}($G$, $D(C, \outX)$, $\outX$) \Comment{Outer variables}
\State $G_W$ = $G \setminus S_O$
\State $V_O$ = $\bigcup\{ Comp \mid Comp \in \textsc{SCC}(G_W), Comp \cap \outX \neq \emptyset\}$
\State $G_O$ = $G[V_O \cup S_O] \cup \textsc{Clique}(S_O)$
\State $TD_O$ = \textsc{TreeDecomposition}($G_O$)
\State $S_{M}$ = \textsc{MinimumSeparator}($G_W$, $D(C, \outX \cup \midX)$, $\midX$) \Comment{Middle variables}
\State $G_W$ = $G_W \setminus S_M$
\State $V_M$ = $\bigcup\{ Comp \mid Comp \in \textsc{SCC}(G_W), Comp \cap \midX \neq \emptyset\}$
\State $G_M$ = $G[V_M \cup S_M \cup S_O] \cup \textsc{Clique}(S_O)  \cup \textsc{Clique}(S_M)$
\State $TD_M$ = \textsc{TreeDecomposition}($G_M$)
\State $G_I$ = $G \setminus (V_O \cup V_M) \cup \textsc{Clique}(S_O) \cup \textsc{Clique}(S_M)$ \Comment{Inner variables}
\State $TD_I$ = \textsc{TreeDecomposition}($G_I$)
\State \Return \textsc{Combine}($TD_O$, $TD_M$, $TD_I$, $\outX$, $\midX$)
\end{algorithmic}
 \caption{\textsc{3AMC-Decomposition}$(C,\outX, \midX, \innX)$}
\label{alg:td_from_cnf}
\end{algorithm}

\begin{example}[cont.]
\red{Given CNF $\mathcal{C}_{cur}$ and the partition 
\[
\mathbf{X}_{O} = \{v_1, v_2, v_3, v_4\}, \mathbf{X}_{M} = \{x_1, x_2, x_3\}, \mathbf{X}_{I} = \{y_1, y_2\},
\]
Algorithm~\ref{alg:td_from_cnf} would choose $S_O = \{x_1\}$, and produce $TD_{O}$ as the first two rows of the left tree decomposition in Figure~\ref{fig:primal_td_running}.
$S_M$ would be chosen as $\{x_2, x_3\}$, and the decomposition contain the third bag from the top of the left tree decomposition in Figure~\ref{fig:primal_td_running} as a a singular bag. 
Finally, $TD_I$ could consist of the bottom four rows.
Combining them would result in the left decomposition in Figure~\ref{fig:primal_td_running}.}
\end{example}

\red{The last step in our implementation is to evaluate our 3AMC-instances over the constrained circuits produced by c2d~\citep{DBLP:conf/ecai/Darwiche04}.
For this, we use the standard approach of~\cite{DBLP:journals/tplp/KieselTK22}.}

\section{Experiments}
\label{sec:experiments}
We implemented the two aforementioned algorithms in Python.
We integrated the enumeration based algorithm into the open source PASTA solver~\citep{AzzBellRig2022PASTA} that leverages clingo~\citep{gebser2019clingo} to compute the answer sets.\footnote{Available at \url{https://github.com/damianoazzolini/pasta}.}
The algorithm based on 3AMC is built on top of aspmc~\citep{DBLP:conf/kr/EiterHK21} and we call it aspmc3.\footnote{Available at \url{https://github.com/raki123/aspmc}.}
During the discussion of the results, we denote them as PASTA and aspmc3, respectively.
We ran the experiments on a computer with Intel\textsuperscript{\textregistered} Xeon\textsuperscript{\textregistered} E5-2630v3 running at 2.40 GHz with 8 Gb of RAM and a time limit of 8 hours.
Execution times are computed with the bash command \texttt{time} and we report the \texttt{real} field.
We generated six synthetic datasets for the experiments.
In the following examples, we report the aspmc3 version of the code.
The programs are the same for PASTA except for the negation symbol: \red{$not$ for PASTA and $\backslash+$ for aspmc3}.
All the probabilities of the probabilistic facts are \red{randomly} set.
\red{In the following snippets we will use the values 0.1, 0.2, 0.3, and 0.4 for conciseness.}
Moreover, every PASP obtained from every strategy has at least one answer set per world.

As a first test (\textit{t1}), we fix the number $n$ of probabilistic facts to 2, 5, 10, and 15 and increase the number $d$ of decision atoms from 0 until we get a memory error or reach the timeout.
We associate a utility of 2 to $qr$ and -12 to $nqr$, as in Example~\ref{ex:running_dummy}.
We use $da/1$ for decision atoms and $a/1$ for probabilistic facts.
We identify the different individuals of the programs with increasing integers, starting from 0, and, in each of these, we add a rule 
$qr \impl a(j), da(i)$ if $i$ is even and two rules
$qr \impl da(i), a(j), \backslash + nqr$ and
$nqr \impl da(i), a(j), \backslash + qr$,
if $i$ is odd, where $j = i \% n$.
For example, with $n = 2$ and $d = 4$, we have:
\begin{lstlisting}
0.1::a(0). 0.2::a(1).
decision da(0). decision da(1). decision da(2). decision da(3).
utility(qr,2). utility(nqr,-12).
qr:- a(0), da(0). 
qr:- a(0), da(2). 
qr:- a(1), da(1), \+ nqr. 
nqr:- a(1), da(1), \+ qr.
qr:- a(1), da(3), \+ nqr. 
nqr:- a(1), da(3), \+ qr.
\end{lstlisting}

In a second test (\textit{t2}) we consider a dual scenario w.r.t. \textit{t1}: here, we fix the number of decision atoms to 2, 5, 10, and 15, and increase the number of probabilistic facts from 0 until we get a memory error or reach the timeout.
The generation of the rules follows the same pattern of \textit{t1}, but with probabilistic facts swapped with decision atoms.
Similarly, we still consider two utility attributes.
For example, with $n = 4$ and $d = 2$, we have:
\begin{lstlisting}
0.1::a(0). 0.2::a(1). 0.3::a(2). 0.4::a(3).
decision da(0). decision da(1).
utility(qr,2). utility(nqr,-12).
qr:- a(0), da(0).
qr:- a(2), da(0).
qr:- a(1), da(1), \+ nqr.
nqr:- a(1), da(1), \+ qr.
qr:- a(3), da(1), \+ nqr.
nqr:- a(3), da(1), \+ qr.
\end{lstlisting}

In a third test (\textit{t3}), for every index $i$, we insert both a probabilistic fact and a decision atom.
We add a rule 
$qr \impl a(i), da(i)$ if $i$ is even and two rules
$qr \impl a(i), da(i), \backslash + nqr$ and  
$nqr \impl a(i), da(i), \backslash + qr$ if $i$ is odd.
Moreover, we add a rule $rda(i) \impl da(i)$ for every $i$ and associate a random utility between -10 and 10 to each $rda(i)$, in addition to $qr/0$ and $nqr/0$.
For example, with $n = d = 4$ we have:
\begin{lstlisting}
0.1::a(0). 0.2::a(1). 0.3::a(2). 0.4::a(3).
decision da(0). decision da(1).
decision da(2). decision da(3).
utility(qr,2). utility(nqr,-12).
utility(rda(0),-6.48). utility(rda(1),3.44).
utility(rda(2),0.49). utility(rda(3),9.63).
rda(0) :- da(0). rda(1) :- da(1).
rda(2) :- da(2). rda(3) :- da(3).
qr:- a(0), da(0). 
qr:- a(2), da(2).
qr:- a(1), da(1), \+ nqr.
nqr:- a(1), da(1), \+ qr.
qr:- a(3), da(3), \+ nqr.
nqr:- a(3), da(3), \+ qr.
\end{lstlisting}

In a fourth test (\textit{t4}), we adopt the same setting of \textit{t3} but the utilities are given only to $qr/0$ and $nqr/0$.

In another test (\textit{t5}), we consider three rules:
$qr \impl \bigwedge_{i \ even} a(i), da(i)$, 
$qr \impl \bigwedge_{i \ odd} a(i), da(i), \backslash+ nqr$, and 
$nqr \impl \bigwedge_{i \ odd} a(i), da(i), \backslash+ qr$,
and a rule $rda(i) \impl da(i)$ for every $i$.
Utilities are associated with $rda/1$ atoms (a random integer between -10 and 10), $qr/0$ (utility 2), and $nqr/0$ (utility -12).
Here as well we test instances with increasing maximum index until we get a memory error or reach the timeout.
For example, with 4 decision atoms and 4 probabilistic facts we have the following program:

\begin{lstlisting}
0.1::a(0). 0.2::a(1). 0.3::a(2). 0.4::a(3).
decision da(0). decision da(1).
decision da(2). decision da(3).
utility(qr,2). utility(nqr,-12).
utility(rda(0),-3.76). utility(rda(1),-7.22).
utility(rda(2),-0.46). utility(rda(3),3.21).
rda(0) :- da(0). 
rda(1) :- da(1).
rda(2) :- da(2).
rda(3) :- da(3).
qr:- a(0), da(0), a(2), da(2).
qr:- a(1), da(1), a(3), da(3), \+ nqr.
nqr:- a(1), da(1), a(3), da(3), \+ qr.
\end{lstlisting}

In a last test (\textit{t6}), we consider instances of Example~\ref{ex:motivating} with an increasing number of people and products involved.
We associate a probabilistic fact $shops(i)$ and decision $target(i)$ for every individual $i$, where $i$ ranges between 1 and $n$, the size of the instance. 
Then, for each $i$, we add a rule with two atoms in the head that define two different items that a person $i$ can buy, as in Example~\ref{ex:motivating}.
Finally, there is an additional rule $rbi(j) \impl item(j)$ for every item $j$ and an utility attribute for every decision atom and $rbi/1$.
For example, the instance of size 10 has 10 probabilistic facts, 10 decision atoms, 10 rules 
$buy(item(a),i) ; buy(item(b),i) \impl target(i), shops(i)$
where $i$ is the person and $item(a)$ and $item(b)$ represent two different products, randomly sampled from a list of 100 different products, 10 utility attributes of the form $utility(target(i),vt)$, one for every person $i$ where $vt$ is a random integer between -5 and 5,
a rule $rb(j) \impl buy(item(j),i)$ for every selected product ($2 \cdot 10$ in total) $item(j)$ and person $i$,
and one utility attribute $utility(rb(j),vu)$ for every such rule where $vu$ is a random integer between -10 and 10.

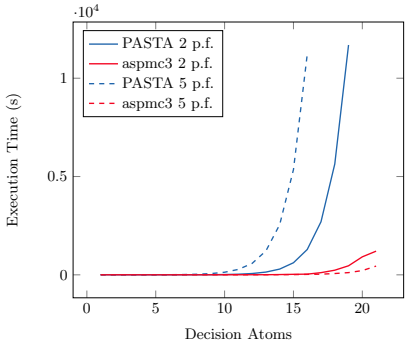
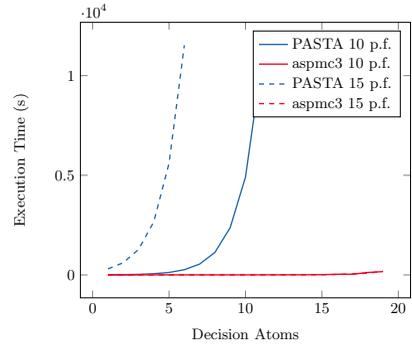
\begin{figure}[t]
  \centering
  \begin{subfigure}{\textwidthsubfigure\textwidth}
  \centering
  \resizebox{\textwidthresizebox\textwidth}{!}{%
  \begin{tikzpicture}
  \begin{axis}[
      xlabel={Decision Atoms},
      ylabel={Execution Time (s)},
      legend pos=north west,
      grid style=dashed,
      legend cell align={left},every axis plot/.append style={thick},
      ytick={0,5000,10000}]
  
      \addplot[color = dark_blue_plot]
      coordinates {
        (1,5.533)
        (2,1.294)
        (3,1.262)
        (4,1.336)
        (5,1.502)
        (6,1.883)
        (7,2.653)
        (8,4.386)
        (9,7.995)
        (10,15.849)
        (11,32.277)
        (12,67.209)
        (13,140.031)
        (14,292.78499999999997)
        (15,614.701)
        (16,1287.899)
        (17,2700.348)
        (18,5636.393)
        (19,11690.148)
      };
      \addlegendentry{PASTA 2 p.f.}
  
      \addplot[color = red_plot] 
      coordinates {
        (1,2.824)
        (2,3.164)
        (3,3.034)
        (4,3.158)
        (5,3.058)
        (6,3.085)
        (7,3.059)
        (8,3.275)
        (9,3.418)
        (10,3.707)
        (11,3.963)
        (12,5.559)
        (13,6.523)
        (14,14.254)
        (15,26.19)
        (16,35.473)
        (17,111.445)
        (18,235.512)
        (19,460.699)
        (20,915.548)
        (21,1204.535)
      };
      \addlegendentry{aspmc3 2 p.f.}
  
      \addplot[color = dark_blue_plot, dashed] 
      coordinates {
        (1,5.986)
        (2,1.518)
        (3,1.826)
        (4,2.554)
        (5,4.097)
        (6,7.445)
        (7,14.78)
        (8,30.104)
        (9,62.909)
        (10,132.376)
        (11,276.631)
        (12,582.013)
        (13,1216.679)
        (14,2553.019)
        (15,5351.596)
        (16,11205.107)
      };
      \addlegendentry{PASTA 5 p.f.}
  
      \addplot[color = red_plot, dashed] 
      coordinates {
        (1,2.597)
        (2,3.174)
        (3,4.042)
        (4,3.025)
        (5,4.015)
        (6,3.032)
        (7,4.06)
        (8,3.392)
        (9,3.302)
        (10,3.57)
        (11,3.855)
        (12,4.406)
        (13,5.747)
        (14,8.411)
        (15,11.027)
        (16,23.365)
        (17,33.03)
        (18,67.086)
        (19,113.499)
        (20,216.752)
        (21,444.349)
      };
      \addlegendentry{aspmc3 5 p.f.}
  
  \end{axis}
  \end{tikzpicture}
  }
  \caption{2 and 5 probabilistic facts.}
  \label{subfig:t1_2_5_pf}
  \end{subfigure}%
  \hfill
  \begin{subfigure}{\textwidthsubfigure\textwidth}
  \centering
  \resizebox{\textwidthresizebox\textwidth}{!}{%
  \begin{tikzpicture}
  \begin{axis}[
      xlabel={Decision Atoms},
      ylabel={Execution Time (s)},
      legend pos=north east,
      grid style=dashed,
      legend cell align={left},every axis plot/.append style={thick},
      ytick={0,5000,10000}]
  
      \addplot[color = dark_blue_plot] 
        coordinates {
          (1,11.701)
          (2,14.631)
          (3,29.514)
          (4,61.506)
          (5,125.376)
          (6,261.911)
          (7,546.115)
          (8,1128.908)
          (9,2374.951)
          (10,4907.726)
          (11,10318.019)
        };
      \addlegendentry{PASTA 10 p.f.}
  
      \addplot[color = red_plot] 
        coordinates {
          (1,2.532)
          (2,3.081)
          (3,4.108)
          (4,4.067)
          (5,3.229)
          (6,4.143)
          (7,4.296)
          (8,4.281)
          (9,4.306)
          (10,4.362)
          (11,4.696)
          (12,5.45)
          (13,6.246)
          (14,8.737)
          (15,13.276)
          (16,34.186)
          (17,36.927)
          (18,123.454)
          (19,168.455)
        };
      \addlegendentry{aspmc3 10 p.f.}
  
      \addplot[color = dark_blue_plot, dashed] 
      coordinates {
        (1,301.046)
        (2,613.868)
        (3,1273.941)
        (4,2663.851)
        (5,5603.538)
        (6,11520.241)
      };
      \addlegendentry{PASTA 15 p.f.}
  
      \addplot[color = red_plot, dashed] 
      coordinates {
        (1,3.439)
        (2,4.197)
        (3,4.522)
        (4,4.188)
        (5,4.884)
        (6,4.321)
        (7,6.332)
        (8,4.678)
        (9,5.379)
        (10,4.647)
        (11,9.525)
        (12,5.746)
        (13,10.616)
        (14,10.102)
        (15,14.838)
        (16,25.737)
        (17,49.794)
        (18,90.074)
        (19,182.89)
      };
      \addlegendentry{aspmc3 15 p.f.}
  \end{axis}
  \end{tikzpicture}
  }
  \caption{10 and 15 probabilistic facts.}
  \label{subfig:t1_10_15_pf}
  \end{subfigure}%
  \caption{Execution times for PASTA and aspmc3 in $t1$ with a fixed number of probabilistic facts and an increasing number of decision atoms.}
  \label{fig:t1_pf}
  \end{figure}

\begin{figure}[t]
\centering
\begin{subfigure}{\textwidthsubfigure\textwidth}
\centering
\resizebox{\textwidthresizebox\textwidth}{!}{%
\begin{tikzpicture}
\begin{axis}[
    xlabel={Probabilistic Facts},
    ylabel={Execution Time (s)},
    legend pos=north west,
    grid style=dashed,
    legend cell align={left},every axis plot/.append style={thick},
    ytick={0,5000,10000}]

    \addplot[color = dark_blue_plot] 
        coordinates {
          (1,4.805)
          (2,1.265)
          (3,1.258)
          (4,1.335)
          (5,1.491)
          (6,1.852)
          (7,2.563)
          (8,4.151)
          (9,7.528)
          (10,14.666)
          (11,32.638)
          (12,67.461)
          (13,140.102)
          (14,293.14300000000003)
          (15,614.163)
          (16,1281.319)
          (17,2675.978)
          (18,5592.884)
          (19,11685.054)          
        };
    \addlegendentry{PASTA 2 d.a.}

    \addplot[color = red_plot] 
        coordinates {
          (1,3.184)
          (2,3.088)
          (3,4.04)
          (4,4.009)
          (5,3.04)
          (6,3.026)
          (7,3.046)
          (8,3.051)
          (9,4.043)
          (10,3.168)
          (11,4.11)
          (12,4.089)
          (13,4.072)
          (14,3.199)
          (15,4.265)
          (16,4.359)
          (17,4.434)
          (18,4.624)
          (19,6.098)
          (20,5.119)
          (21,7.16)
          (22,7.09)
          (23,8.289)
          (24,7.702)
          (25,11.611)
          (26,11.254)
          (27,17.01)
          (28,34.047)
          (29,44.223)
        };
    \addlegendentry{aspmc3 2 d.a.}

    \addplot[color = dark_blue_plot, dashed] 
      coordinates {
        (1,5.223)
        (2,1.531)
        (3,1.842)
        (4,2.568)
        (5,4.096)
        (6,7.418)
        (7,14.345)
        (8,29.196)
        (9,60.484)
        (10,127.806)
        (11,288.738)
        (12,607.815)
        (13,1263.873)
        (14,2643.308)
        (15,5533.241)
        (16,11447.358)        
      };

    \addlegendentry{PASTA 5 d.a.}

    \addplot[color = red_plot, dashed] 
    coordinates {
      (1,3.165)
      (2,3.052)
      (3,3.044)
      (4,3.046)
      (5,4.054)
      (6,3.095)
      (7,4.177)
      (8,4.102)
      (9,4.063)
      (10,3.223)
      (11,3.186)
      (12,3.423)
      (13,4.369)
      (14,4.902)
      (15,4.944)
      (16,5.508)
      (17,5.192)
      (18,8.103)
      (19,6.604)
      (20,6.144)
      (21,10.029)
      (22,8.369)
      (23,8.792)
      (24,30.044)
      (25,16.332)
      (26,32.106)
      (27,21.589)
      (28,60.498)
      (29,92.175)
    };
    \addlegendentry{aspmc3 5 d.a.}

\end{axis}
\end{tikzpicture}
}
\caption{2 and 5 decision atoms.}
\label{subfig:t2_2_5_da}
\end{subfigure}%
\hfill
\begin{subfigure}{\textwidthsubfigure\textwidth}
\centering
\resizebox{\textwidthresizebox\textwidth}{!}{%
\begin{tikzpicture}
\begin{axis}[
    xlabel={Probabilistic Facts},
    ylabel={Execution Time (s)},
    legend pos=north east,
    grid style=dashed,
    legend cell align={left},every axis plot/.append style={thick}, ytick={0,5000,10000}]
    
    \addplot[color = dark_blue_plot] 
      coordinates {
        (1,14.274)
        (2,15.564)
        (3,31.043)
        (4,64.056)
        (5,131.735)
        (6,271.631)
        (7,560.373)
        (8,1151.687)
        (9,2367.957)
        (10,4791.554)
        (11,11019.089)        
      };
    \addlegendentry{PASTA 10 d.a.}

    \addplot[color = red_plot] 
      coordinates {
        (1,4.186)
        (2,3.565)
        (3,3.342)
        (4,3.259)
        (5,3.452)
        (6,4.455)
        (7,4.33)
        (8,4.276)
        (9,4.296)
        (10,4.28)
        (11,4.227)
        (12,4.78)
        (13,4.783)
        (14,5.517)
        (15,4.483)
        (16,5.376)
        (17,4.898)
        (18,8.24)
        (19,5.227)
        (20,26.539)
        (21,8.466)
        (22,19.987)
        (23,9.206)
        (24,52.476)
        (25,11.991)
        (26,43.582)
        (27,34.238)
        (28,218.191)
        (29,81.741)
      };
    \addlegendentry{aspmc3 10 d.a.}

    \addplot[color = dark_blue_plot, dashed] 
      coordinates {
        (1,302.167)
        (2,613.054)
        (3,1272.697)
        (4,2585.166)
        (5,5229.528)
        (6,10768.5)        
      };
    \addlegendentry{PASTA 15 d.a.}

    \addplot[color = red_plot, dashed] 
      coordinates {
        (1,36.966)
        (2,23.977)
        (3,9.385)
        (4,10.068)
        (5,10.536)
        (6,15.882)
        (7,13.888)
        (8,12.842)
        (9,17.317)
        (10,11.992)
        (11,14.521)
        (12,11.964)
        (13,15.317)
        (14,12.643)
        (15,14.322)
        (16,14.686)
        (17,18.743)
        (18,58.419)
        (19,61.385)
        (20,42.808)
        (21,44.663)
        (22,105.857)
        (23,116.432)
        (24,179.357)
        (25,308.662)
        (26,1410.036)
      };
    \addlegendentry{aspmc3 15 d.a.}

\end{axis}
\end{tikzpicture}
}
\caption{10 and 15 decision atoms.}
\label{subfig:t2_10_15_da}
\end{subfigure}%
\caption{Execution times for PASTA and aspmc3 in $t2$ with a fixed number of decision atoms and an increasing number of probabilistic facts.}
\label{fig:t2_da}
\end{figure}
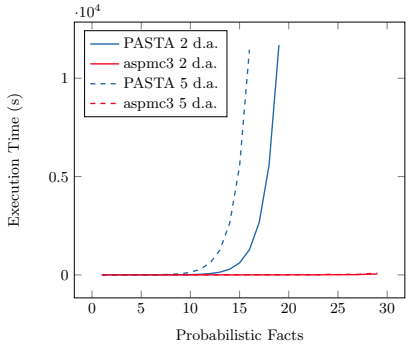
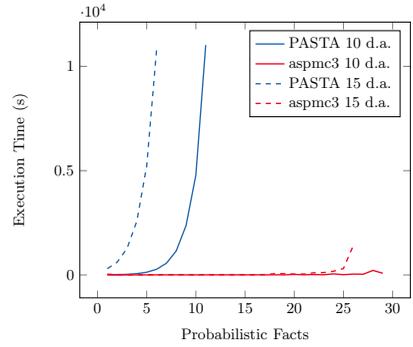

\begin{figure}[t]
\centering
\begin{subfigure}{\textwidthsubfigure\textwidth}
\centering
\resizebox{\textwidthresizebox\textwidth}{!}{%
\begin{tikzpicture}
\begin{axis}[
    xlabel={Decision Atoms},
    ylabel={Execution Time (s)},
    legend pos=north west,
    grid style=dashed,
    legend cell align={left},every axis plot/.append style={thick},
    y tick label style={/pgf/number format/fixed, /pgf/number format/.cd, /tikz/.cd}]

    \addplot[color = dark_blue_plot] 
      coordinates {
        (2,1.334)
        (3,1.374)
        (4,1.946)
        (5,4.855)
        (6,18.032)
        (7,77.883)
        (8,343.656)
        (9,1664.92)
        (10,6900.373)
      };
    \addlegendentry{$t3$ PASTA}

    \addplot[color = red_plot] 
      coordinates {
        (1,3.141)
        (2,2.94)
        (3,3.107)
        (4,3.982)
        (5,3.966)
        (6,4.153)
        (7,4.028)
        (8,4.126)
        (9,4.289)
        (10,4.687)
        (11,5.535)
        (12,6.976)
        (13,10.358)
        (14,14.078)
        (15,25.951)
        (16,59.59)
        (17,75.13)
        (18,206.855)
      };
    \addlegendentry{$t3$ aspmc3}

    \addplot[color = dark_blue_plot, dashed] 
      coordinates {
        (1,6.85)
        (2,1.324)
        (3,1.809)
        (4,4.009)
        (5,13.931)
        (6,59.132)
        (7,258.48)
        (8,1125.213)
        (9,4823.093)        
      };
    \addlegendentry{$t4$ PASTA}

    \addplot[color = red_plot, dashed] 
      coordinates {
        (1,3.076)
        (2,2.954)
        (3,2.968)
        (4,3.979)
        (5,3.018)
        (6,4.061)
        (7,4.082)
        (8,4.352)
        (9,4.36)
        (10,4.539)
        (11,5.2)
        (12,7.13)
        (13,8.379)
        (14,14.181)
        (15,25.117)
        (16,54.992)
        (17,86.121)
        (18,177.461)
      };
    \addlegendentry{$t4$ aspmc3}

\end{axis}
\end{tikzpicture}
}
\caption{$t3$ and $t4$.}
\label{subfig:t3_t4}
\end{subfigure}%
\hfill
\begin{subfigure}{\textwidthsubfigure\textwidth}
\centering
\resizebox{\textwidthresizebox\textwidth}{!}{%
\begin{tikzpicture}
\begin{axis}[
    xlabel={Decision Atoms},
    ylabel={Execution Time (s)},
    legend pos=north east,
    grid style=dashed,
    legend cell align={left},every axis plot/.append style={thick},
    y tick label style={/pgf/number format/fixed, /pgf/number format/.cd,fixed zerofill, /tikz/.cd}]

    \addplot[color = dark_blue_plot] 
      coordinates {
        (1,6.559)
        (2,1.271)
        (3,1.346)
        (4,1.945)
        (5,4.791)
        (6,18.084)
        (7,76.867)
        (8,337.582)
        (9,1643.685)
        (10,7036.105)        
      };
    \addlegendentry{$t5$ PASTA}

    \addplot[color = red_plot] 
      coordinates {
        (1,0.999)
        (2,3.027)
        (3,4.013)
        (4,4.004)
        (5,4.037)
        (6,5.028)
        (7,4.047)
        (8,4.042)
        (9,4.055)
        (10,4.104)
        (11,5.165)
        (12,5.122)
        (13,5.094)
        (14,5.181)
        (15,5.106)
        (16,4.267)
        (17,5.174)
        (18,5.179)
        (19,5.283)
        (20,4.415)
      };
    \addlegendentry{$t5$ aspmc3}

    \addplot[color = dark_blue_plot, dashed]
      coordinates {
        (1,6.307)
        (2,1.224)
        (3,1.206)
        (4,2.481)
        (5,7.33)
        (6,26.773)
        (7,128.17)
        (8,562.835)
        (9,2521.462)
        (10,11087.993)
      };
    \addlegendentry{$t6$ PASTA}

    \addplot[color = red_plot, dashed]
      coordinates {
        (1,3.163)
        (2,3.074)
        (3,3.104)
        (4,4.119)
        (5,3.271)
        (6,3.476)
        (7,3.947)
        (8,5.26)
        (9,7.099)
        (10,11.522)
        (11,12.848)
        (12,40.98)
        (13,82.586)
        (14,169.897)
        (15,361.681)
      };
    \addlegendentry{$t6$ aspmc3}

\end{axis}
\end{tikzpicture}
}
\caption{$t5$ and $t6$.}
\label{subfig:t5_t6}
\end{subfigure}%
\caption{Execution times for PASTA and aspmc3 in $t3$, $t4$, $t5$, and $t6$.}
\label{fig:t3_t4_t5_t6}
\end{figure}
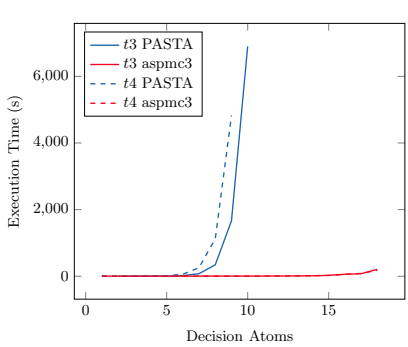
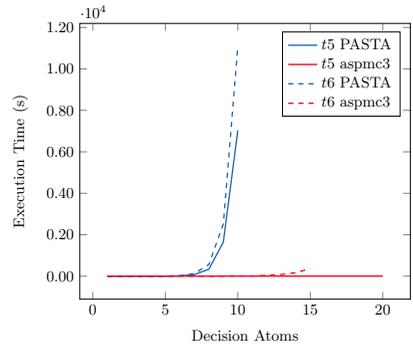

Figures~\ref{fig:t1_pf},~\ref{fig:t2_da}, and~\ref{fig:t3_t4_t5_t6} show the execution times of the six tests: PASTA cannot scale over programs with more than 20 decision atoms and probabilistic facts combined.
aspmc3 can manage larger instances in terms of decision atoms and probabilistic facts in a fraction of time w.r.t. PASTA.

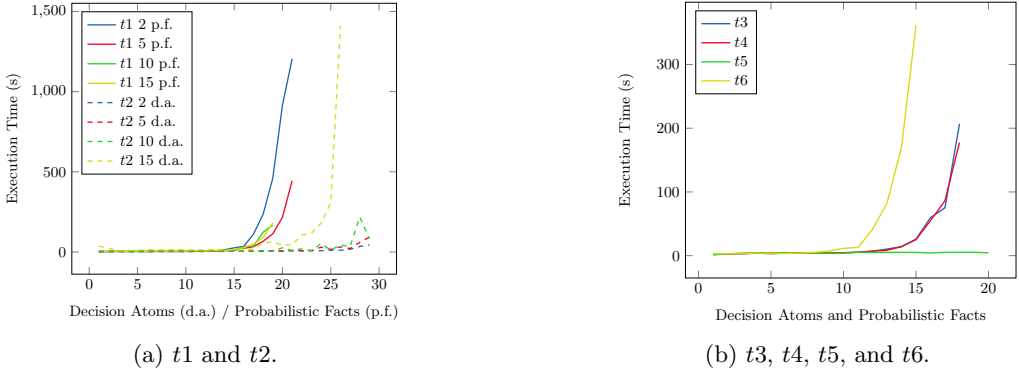
\begin{figure}[t]
\centering
\begin{subfigure}{\textwidthsubfigure\textwidth}
\centering
\resizebox{\textwidthresizebox\textwidth}{!}{%
\begin{tikzpicture}
\begin{axis}[
    xlabel={Decision Atoms (d.a.) / Probabilistic Facts (p.f.)},
    ylabel={Execution Time (s)},
    legend pos=north west,
    grid style=dashed,
    legend cell align={left},every axis plot/.append style={thick},
    y tick label style={/pgf/number format/fixed, /pgf/number format/precision=0,/pgf/number format/.cd,fixed zerofill, /tikz/.cd}]

    \addplot[color = dark_blue_plot]
      coordinates {
        (1,2.824)
        (2,3.164)
        (3,3.034)
        (4,3.158)
        (5,3.058)
        (6,3.085)
        (7,3.059)
        (8,3.275)
        (9,3.418)
        (10,3.707)
        (11,3.963)
        (12,5.559)
        (13,6.523)
        (14,14.254)
        (15,26.19)
        (16,35.473)
        (17,111.445)
        (18,235.512)
        (19,460.699)
        (20,915.548)
        (21,1204.535)
      };

    \addlegendentry{$t1$ 2 p.f.}

    \addplot[color = red_plot] 
      coordinates {
        (1,2.597)
        (2,3.174)
        (3,4.042)
        (4,3.025)
        (5,4.015)
        (6,3.032)
        (7,4.06)
        (8,3.392)
        (9,3.302)
        (10,3.57)
        (11,3.855)
        (12,4.406)
        (13,5.747)
        (14,8.411)
        (15,11.027)
        (16,23.365)
        (17,33.03)
        (18,67.086)
        (19,113.499)
        (20,216.752)
        (21,444.349)
      };
    \addlegendentry{$t1$ 5 p.f.}

    \addplot[color = green_plot] 
        coordinates {
          (1,2.532)
          (2,3.081)
          (3,4.108)
          (4,4.067)
          (5,3.229)
          (6,4.143)
          (7,4.296)
          (8,4.281)
          (9,4.306)
          (10,4.362)
          (11,4.696)
          (12,5.45)
          (13,6.246)
          (14,8.737)
          (15,13.276)
          (16,34.186)
          (17,36.927)
          (18,123.454)
          (19,168.455)
        };
    \addlegendentry{$t1$ 10 p.f.}

    \addplot[color = yellow_plot] 
      coordinates {
        (1,3.439)
        (2,4.197)
        (3,4.522)
        (4,4.188)
        (5,4.884)
        (6,4.321)
        (7,6.332)
        (8,4.678)
        (9,5.379)
        (10,4.647)
        (11,9.525)
        (12,5.746)
        (13,10.616)
        (14,10.102)
        (15,14.838)
        (16,25.737)
        (17,49.794)
        (18,90.074)
        (19,182.89)
      };
    \addlegendentry{$t1$ 15 p.f.}


    \addplot[color = dark_blue_plot, dashed] 
      coordinates {
        (1,3.184)
        (2,3.088)
        (3,4.04)
        (4,4.009)
        (5,3.04)
        (6,3.026)
        (7,3.046)
        (8,3.051)
        (9,4.043)
        (10,3.168)
        (11,4.11)
        (12,4.089)
        (13,4.072)
        (14,3.199)
        (15,4.265)
        (16,4.359)
        (17,4.434)
        (18,4.624)
        (19,6.098)
        (20,5.119)
        (21,7.16)
        (22,7.09)
        (23,8.289)
        (24,7.702)
        (25,11.611)
        (26,11.254)
        (27,17.01)
        (28,34.047)
        (29,44.223)
      };
    \addlegendentry{$t2$ 2 d.a.}

    \addplot[color = red_plot, dashed] 
        coordinates {
          (1,3.165)
          (2,3.052)
          (3,3.044)
          (4,3.046)
          (5,4.054)
          (6,3.095)
          (7,4.177)
          (8,4.102)
          (9,4.063)
          (10,3.223)
          (11,3.186)
          (12,3.423)
          (13,4.369)
          (14,4.902)
          (15,4.944)
          (16,5.508)
          (17,5.192)
          (18,8.103)
          (19,6.604)
          (20,6.144)
          (21,10.029)
          (22,8.369)
          (23,8.792)
          (24,30.044)
          (25,16.332)
          (26,32.106)
          (27,21.589)
          (28,60.498)
          (29,92.175)
        };
    \addlegendentry{$t2$ 5 d.a.}

    \addplot[color = green_plot, dashed] 
        coordinates {
          (1,4.186)
          (2,3.565)
          (3,3.342)
          (4,3.259)
          (5,3.452)
          (6,4.455)
          (7,4.33)
          (8,4.276)
          (9,4.296)
          (10,4.28)
          (11,4.227)
          (12,4.78)
          (13,4.783)
          (14,5.517)
          (15,4.483)
          (16,5.376)
          (17,4.898)
          (18,8.24)
          (19,5.227)
          (20,26.539)
          (21,8.466)
          (22,19.987)
          (23,9.206)
          (24,52.476)
          (25,11.991)
          (26,43.582)
          (27,34.238)
          (28,218.191)
          (29,81.741)
        };
    \addlegendentry{$t2$ 10 d.a.}

    \addplot[color = yellow_plot, dashed] 
        coordinates {
          (1,36.966)
          (2,23.977)
          (3,9.385)
          (4,10.068)
          (5,10.536)
          (6,15.882)
          (7,13.888)
          (8,12.842)
          (9,17.317)
          (10,11.992)
          (11,14.521)
          (12,11.964)
          (13,15.317)
          (14,12.643)
          (15,14.322)
          (16,14.686)
          (17,18.743)
          (18,58.419)
          (19,61.385)
          (20,42.808)
          (21,44.663)
          (22,105.857)
          (23,116.432)
          (24,179.357)
          (25,308.662)
          (26,1410.036)
        };
    \addlegendentry{$t2$ 15 d.a.}

\end{axis}
\end{tikzpicture}
}
\caption{$t1$ and $t2$.}
\label{subfig:aspmc_t1_t2}
\end{subfigure}%
\hfill
\begin{subfigure}{\textwidthsubfigure\textwidth}
\centering
\resizebox{\textwidthresizebox\textwidth}{!}{%
\begin{tikzpicture}
\begin{axis}[
    xlabel={Decision Atoms and Probabilistic Facts},
    ylabel={Execution Time (s)},
    legend pos=north west,
    grid style=dashed,
    legend cell align={left},every axis plot/.append style={thick},
    y tick label style={/pgf/number format/fixed, /pgf/number format/precision=0,/pgf/number format/.cd,fixed zerofill, /tikz/.cd}]

    \addplot[color = dark_blue_plot] 
        coordinates {
          (1,3.141)
          (2,2.94)
          (3,3.107)
          (4,3.982)
          (5,3.966)
          (6,4.153)
          (7,4.028)
          (8,4.126)
          (9,4.289)
          (10,4.687)
          (11,5.535)
          (12,6.976)
          (13,10.358)
          (14,14.078)
          (15,25.951)
          (16,59.59)
          (17,75.13)
          (18,206.855)
        };
    \addlegendentry{$t3$}

    \addplot[color = red_plot] 
        coordinates {
          (1,3.076)
          (2,2.954)
          (3,2.968)
          (4,3.979)
          (5,3.018)
          (6,4.061)
          (7,4.082)
          (8,4.352)
          (9,4.36)
          (10,4.539)
          (11,5.2)
          (12,7.13)
          (13,8.379)
          (14,14.181)
          (15,25.117)
          (16,54.992)
          (17,86.121)
          (18,177.461)
        };
    \addlegendentry{$t4$}

    \addplot[color = green_plot] 
        coordinates {
          (1,0.999)
          (2,3.027)
          (3,4.013)
          (4,4.004)
          (5,4.037)
          (6,5.028)
          (7,4.047)
          (8,4.042)
          (9,4.055)
          (10,4.104)
          (11,5.165)
          (12,5.122)
          (13,5.094)
          (14,5.181)
          (15,5.106)
          (16,4.267)
          (17,5.174)
          (18,5.179)
          (19,5.283)
          (20,4.415)
        };
    \addlegendentry{$t5$}

    \addplot[color = yellow_plot] 
        coordinates {
          (1,3.163)
          (2,3.074)
          (3,3.104)
          (4,4.119)
          (5,3.271)
          (6,3.476)
          (7,3.947)
          (8,5.26)
          (9,7.099)
          (10,11.522)
          (11,12.848)
          (12,40.98)
          (13,82.586)
          (14,169.897)
          (15,361.681)
        };
    \addlegendentry{$t6$}

\end{axis}
\end{tikzpicture}
}
\caption{$t3$, $t4$, $t5$, and $t6$.}
\label{subfig:aspmc_t3_t4_t5_t6}
\end{subfigure}%
\caption{Results for aspmc3 in the six tests.}
\label{fig:aspmc}
\end{figure}

To better assess the performance of aspmc3, we grouped and reported the results also in Figure~\ref{fig:aspmc}.
From Figure~\ref{subfig:aspmc_t1_t2} we can see that the instances of $t1$ take more time than those of $t2$ with the same index, so the number of decision atoms impacts more than the number of probabilistic facts (being their total number the same) on the execution time.
For $t3$ and $t4$, the execution times are similar, reaching the memory limit at instance 19.
Nevertheless, for size 18, aspmc3 took approximately 200 seconds (Figure~\ref{subfig:aspmc_t3_t4_t5_t6}) to complete: this shows that the program is able to manage up to 18 decision facts and probabilistic facts relatively quickly.
For $t5$, the execution time is almost constant: this is due to the knowledge compilation step, since the program is composed of only three rules with bodies with increasing length.
For $t6$, aspmc3 reaches the memory limit at instance 16.
For the instance of size 15, the execution time is approximately 360 seconds.
Table~\ref{tab:datasets} lists all the tests with the number of decision atoms, probabilistic facts, and utility attributes for the corresponding largest solvable instance.
Overall, the adoption of 3AMC makes a huge improvement over naive answer set enumeration, even if it requires a non-negligible amount of memory.

\begin{table}[t]
\centering
\caption{Largest solvable instances for each test for aspmc3. In the first column for $t1$ and $t2$ we inserted the number of probabilistic facts (p.f.) and decision atoms (d.a.) between parentheses to indicate the test with that number of p.f. and d.a. fixed.}
\begin{tabular}{||c | c | c | c ||}
Test & \# decision atoms \ & \# probabilistic facts \ & \# utility attributes \
\midline
$t_1$ (2 p.f.) & 21 & 2 & 2 \\
$t_1$ (5 p.f.) & 21 & 5 & 2 \\
$t_1$ (10 p.f.) & 19 & 10 & 2 \\
$t_1$ (15 p.f.) & 19 & 15 & 2 \\
$t_2$ (2 d.a.) & 2 & 29 & 2 \\
$t_2$ (5 d.a.) & 5 & 29 & 2 \\
$t_2$ (10 d.a.) & 10 & 29 & 2 \\
$t_2$ (15 d.a.) & 15 & 26 & 2 \\
$t_3$ & 18 & 18 & 20 \\
$t_4$ & 18 & 18 & 2 \\
$t_5$ & 91 & 91 & 93 \\
$t_6$ & 15 & 15 & 45 \\
\end{tabular}
\label{tab:datasets}
\end{table}

\section{Related Work}
\label{sec:related}
This work is inspired to DTProbLog~\citep{DBLP:conf/aaai/BroeckTOR10}.
If we only consider normal rules, the decision theory task can be expressed with both DTProbLog and our framework, but our framework is more general, since it admits a large subset of the whole ASP syntax.

The possibility of expressing decision theory problems with ASP gathered a lot of research interest in the past years.
The author of~\citep{DBLP:conf/aaai/Brewka02} extended ASP by introducing Logic Programming with Ordered Disjunction (LPODs) based on the use of a new connective called \textit{ordered disjunction} that specifies an order of preferences among the possible answer sets.
This was the starting point for several works:~\cite{brewka2002decision} proposes a framework for quantitative decision making while~\cite{10.1007/978-3-642-22152-1_41} adopt a possibilistic extension of LPODs.
Also~\cite{10.1007/978-3-540-30106-6_49} adopts LPODs and casts the decision theory problem as a constraint satisfaction problem with the goal of identifying the preferred stable models.
Differently from these works, we define uncertainty using probabilistic facts, possible actions using decision atoms, and associate weights (utilities) to (some of the) possible atoms.
Moreover, we do not define preferences over the answer sets, rather over the possible combinations of decision atoms (strategies) and combinations of probabilistic facts (worlds).
Preferences among the possible answer sets can be encoded through weak constraints~\citep{buccafurri2000constraints,calimeri2012asp} which are a standard feature of ASP systems.
$\lpmln$~\citep{DBLP:conf/kr/LeeW16} and P-log~\citep{DBLP:journals/tplp/BaralGR09} are two languages to represent uncertainty with ASP, whose relationship, also with weak constraints, has been extensively explored in~\citep{DBLP:conf/ijcai/BalaiG16,DBLP:conf/aaai/LeeY17}.
The former associates weights with atoms and rules while the latter adopts the so-called random selection rules.
$\lpmln$ supports the computation of the preferred stable models, according to a weight obtained by considering the weighted atoms present in an answer set.
However, it does not consider the decision theory task.
Similar considerations hold for P-log.

We discuss normalization to handle programs where some worlds have no stable models, which has already been proved effective in related scenarios~\citep{Fierens12}.
There are other possible solutions that can be adopted in case of inconsistent programs, such as trying to ``repair'' them~\citep{10.1007/978-3-642-15918-3_9,10.5555/3032027.3032070}.
However, these are often tailored to the considered program.
The authors of~\citep{DBLP:conf/kr/RochaC22} propose the adoption of least undefined stable models, i.e., partial models where the number of undefined atoms is minimum.
With this, it is still possible to assign a semantics to programs where some worlds are inconsistent, and the probability of a query is still defined in terms of probability bounds (lower and upper).
However, they do not discuss and develop an inference algorithm.
Another possibility is the one proposed by~\citep{totis_de_raedt_kimmig_2023}: the authors present smProbLog, an extension of the ProbLog2 framework~\citep{dries2015problog2} that allows worlds to have more than one model and even no models.
Their approach consists in uniformly distributing the probability of a world into its stable models, and to consider three truth values for an atom (true, false, and inconsistent), with three associated probabilities.
In this way, every atom has a sharp probability value.
They also propose a practical algorithm to perform inference.
In general, these two approaches do not consider the whole ASP syntax.

\red{We propose how to find the strategy that maximizes the lower expected reward and the one that maximizes the upper expected reward.
More generally,~\cite{augustin2014introduction} distinguish between non-sequential and sequential decision problems.
In non-sequential problems, the subject must choose only \emph{one} of a number of possible actions, each of which leads to an uncertain reward.
If a subject can express her beliefs through a probability measure, then a common solution is for her to choose the act that \emph{maximizes} her expected utility.
However, there are occasions when information cannot be represented through a linear prevision, but instead by a more general uncertainty model.
In such circumstances, we would like to identify a best act in any choice.
For example, suppose that one must choose between two gambles. 
To identify the preferred gamble from a particular set, we can specify gambles that we do not want to choose.
The optimal set of gambles is what remains after all unacceptable gambles are eliminated. 
In sequential decision problems, the subject may have to make more than one decision, at different times: this problem is displayed on a decision tree.
Branches emerging from square nodes represent acts, and branches emerging from circular nodes represent events.
At the end of each path is a number representing the utility reward of that particular combination of acts and events.
}

\red{Markov Decision Processes (MDP) represent a class of sequential decision-making problems in a stochastic environment.
A planning agent has to deliberate over his/her model of the world to choose an optimal action in each decision stage in order to maximize his/her accumulated reward (or minimize the accumulated cost) given the immediate and long-term uncertain effects of available actions.
There are situations in which it is not easy (or even possible) to define a precise probability measure for a given transition.
In this case, it is necessary to consider a more general version of an MDP known as Markov Decision Processes with Imprecise Probabilities~\citep{pmlr-v62-bueno17a}: in this model, the probability parameters are imprecise and therefore the transition model cannot be specified by a single conditional distribution, but it must be defined by sets of probabilities for each state transition, which are referred to as transition credal sets.
\cite{pmlr-v62-bueno17a} propose a novel language based on Probabilistic Logic Programming, enhanced with decision theoretic constructs such as actions, state fluents and utilities.
They consider interval-valued probabilities attached to independent facts, i.e., a fact has to be associated with a probability interval  $[\alpha, \beta] :: p$, where $p$ is an atom and parameters $\alpha$ and $\beta$ are probability bounds such as $0 \leq \alpha \leq \beta \leq 1$.
In the case of $\alpha=\beta$, one has a standard probabilistic fact.
The semantics of a probabilistic logic program with interval-valued facts is the credal set that consists of all probability distributions that satisfy the constraints (that is, whose marginal probabilities for facts lies within given intervals).
In our work, we share the use of the credal semantics; however, probabilistic facts are annotated with a single probability value.
}

\section{Conclusions and Future Works}
\label{sec:conclusions}
In this paper, we discussed how to encode and solve a decision theory task with Probabilistic Answer Set Programming under the credal semantics.
We proposed the class of decision theoretic probabilistic answer set programs, i.e., probabilistic answer set programs extended with decision atoms, representing the possible actions that can be taken, and utility attributes, representing the rewards that can be obtained.
The goal is to find the two sets of decision atoms that yield the highest lower bound and the highest upper bound for the overall utility, respectively.
We developed an algorithm based on three layers of Algebraic Model Counting and knowledge compilation and compared it against a naive algorithm based on answer set enumeration.
Empirical results show that our approach is able to manage instances of non trivial sizes in a reasonable amount of time.
A possible future work consists of proposing a formalization of the decision theory task also for other semantics for probabilistic ASP.
Moreover, we could consider decision problems also in the case that the probabilistic facts are annotated with probability intervals, exploiting the results for inference in~\cite{AzzRig24-UAI-IC}.

\section*{Acknowledgements} 
This work has been partially supported by Spoke 1 ``FutureHPC \& BigData'' of the Italian Research Center on High-Performance Computing, Big Data and Quantum Computing (ICSC) funded by MUR Missione 4 - Next Generation EU (NGEU) and by Partenariato Esteso PE00000013 - ``FAIR - Future Artificial Intelligence Research'' - Spoke 8 ``Pervasive AI'', funded by MUR through PNRR - M4C2 - Investimento 1.3 (Decreto Direttoriale MUR n. 341 of 15th March 2022) under the Next Generation EU (NGEU).
Damiano Azzolini and Fabrizio Riguzzi are members of the Gruppo Nazionale Calcolo Scientifico -- Istituto Nazionale di Alta Matematica (GNCS-INdAM).

\subsection*{Competing Interests}
The authors declare none.

\bibliographystyle{ACM-Reference-Format}
\bibliography{2022pasp_dt_monofile.bbl}

\end{document}